\documentclass[10pt,twocolumn,letterpaper]{article}

\usepackage{cvpr}              

\newcommand{\datasetname}{\mbox{\textsc{EgoPointVQA}}\xspace}

\newcommand{\methodname}{\mbox{\textsc{HINT}}\xspace}
\newcommand{\methodnamefull}{\underline{H}and \underline{In}tent \underline{T}okens\xspace}

\usepackage{booktabs}          
\usepackage{graphicx}          
\usepackage[table]{xcolor}     
\usepackage{pifont}
\newcommand{\cmark}{\color{blue} \ding{51}}%
\newcommand{\xmark}{\color{red} \ding{55}}%
\definecolor{handtoken}{HTML}{7154DB} 
\definecolor{cvprblue}{rgb}{0.21,0.49,0.74}
\newcommand{\refmain}[1]{\textcolor{cvprblue}{#1}}

\usepackage[most]{tcolorbox}
\usepackage{listings}

\definecolor{cvprblue}{rgb}{0.21,0.49,0.74}
\usepackage[pagebackref,breaklinks,colorlinks,allcolors=cvprblue]{hyperref}

\def\paperID{46058} 
\def\confName{CVPR}
\def\confYear{2026}

\usepackage{multirow}
\usepackage{graphicx}
\usepackage{listings}
\usepackage{tcolorbox}
\usepackage{bbding}
\tcbset{
  colback=gray!10,       
  colframe=black!60,     
  boxrule=0.4pt,
  arc=2pt,
  left=4pt, right=4pt, top=4pt, bottom=4pt
}

\usepackage{xcolor}
\definecolor{mygreen}{RGB}{70, 184, 101}
\usepackage{capt-of} 

\newcommand*{\belowrulesepcolor}[1]{%
  \noalign{%
    \kern-\belowrulesep 
    \begingroup 
      \color{#1}%
      \hrule height\belowrulesep 
    \endgroup 
  }%
} 
\newcommand*{\aboverulesepcolor}[1]{%
  \noalign{%
    \begingroup 
      \color{#1}%
      \hrule height\aboverulesep 
    \endgroup 
    \kern-\aboverulesep 
  }%
}

\usepackage{graphicx}

\title{Do You See What I Am Pointing At? \\Gesture-Based Egocentric Video Question Answering}

\author{
Yura Choi, 
Roy Miles, 
Rolandos Alexandros Potamias
\and
Ismail Elezi, 
Jiankang Deng, 
Stefanos Zafeiriou
\and
{\normalfont Imperial College London} \\[0.2em]
{\url{https://yuuraa.github.io/papers/choi2026egovqa}}
}

\begin{document}

\twocolumn[{%
\renewcommand\twocolumn[1][]{#1}%
\maketitle
\vspace{-0.6em}
\includegraphics[width=\linewidth]{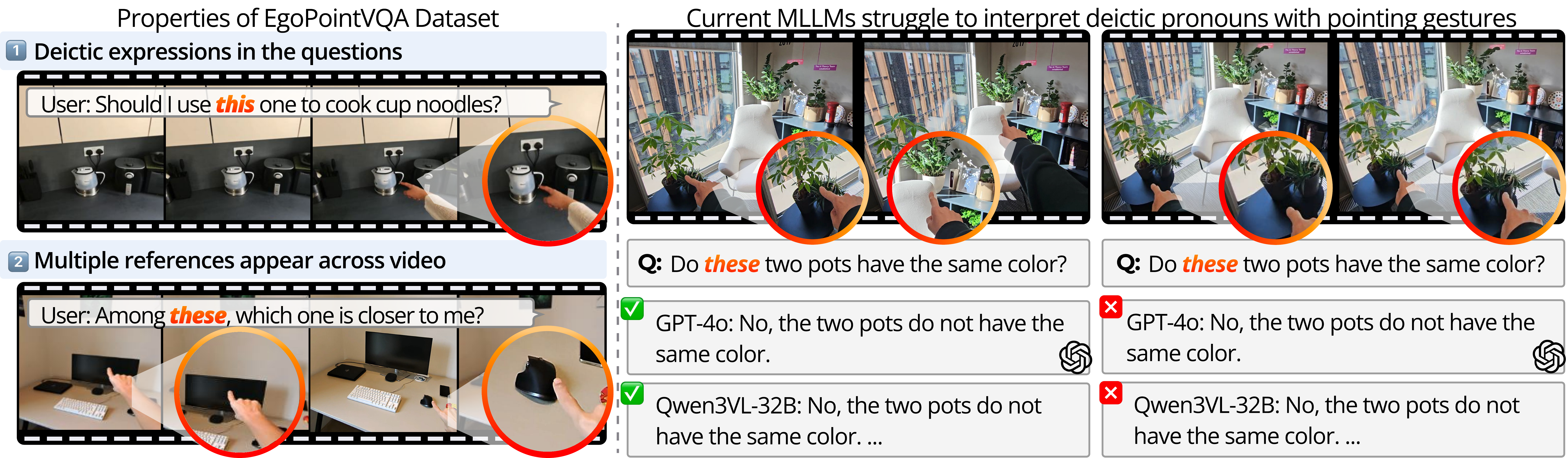}
\vspace{-0.7cm}
\captionof{figure}{
\textbf{Illustration of \datasetname.} 
\textbf{Left}: \datasetname includes questions with deictic pronouns requiring gesture understanding, either identifying single pointed objects (top) or tracking multiple references across frames (bottom).
\textbf{Right}: State-of-the-art models, including GPT-4o~\cite{gpt4o} and Qwen3-VL-32B~\cite{qwen3technicalreport}, fail to resolve the question with pointing gestures, incorrectly stating the two pots have different colors despite both being black. Zoomed circles highlight the pointed objects.
}
\vspace{0.9cm}
\label{fig:fig1_teaser}
}]
\begin{abstract}
\vspace{-1cm}

Understanding and answering questions based on a user’s pointing gesture is essential for next-generation egocentric AI assistants. However, current Multimodal Large Language Models (MLLMs) struggle with such tasks due to the lack of gesture-rich data and their limited ability to infer fine-grained pointing intent from egocentric video.
To address this, we introduce \datasetname, a dataset and benchmark for gesture-grounded egocentric question answering, comprising 4,000 synthetic and 400 real-world videos across multiple deictic reasoning tasks.
Built upon it, we further propose \methodnamefull (HINT), which encodes tokens derived from 3D hand keypoints using an off-the-shelf reconstruction model and interleaves them with the model input to provide explicit spatial and temporal context for interpreting pointing intent.
We show that our model outperforms others in different backbones and model sizes.
In particular, HINT-14B
achieves 68.1\% accuracy, on average over 6 tasks, surpassing the state-of-the-art, InternVL3-14B, by 5.4\%.
To further facilitate the open research, we will release the code, model, and dataset.
\end{abstract}
\vspace{-1em}
\section{Introduction}
\label{sec:intro}
As AI assistants become deeply integrated into daily wearable devices, from augmented and virtual reality (AR/VR) platforms like the Apple Vision Pro and Meta Orion to smart glasses like Meta Ray-Ban~\cite{waisberg2024meta,MetaRayBan2025} with built-in cameras, their ability to understand a user's attention within their environment becomes essential~\cite{lee2024gazepointar,yan2023voilaaaligningvisionlanguagemodels,egogazevqa,rekimoto2025gazellm}. 
In particular, it is important to
resolve spatial references through pointing gestures and deictic expressions (e.g., ``Should I use \textit{this} one?"), which are natural in human communication~\cite{wang2023holoassist,wan2022handmethat,diessel2020demonstratives}.
Thus, an AI assistant must understand not just what it sees, but also where the user is directing their attention.
Typically, this requires: (1) identifying the deictic expression within the question that depends on gesture for meaning, (2) interpreting the hand's movement and pose to understand the user's referential intent, and (3) grounding this intent to identify the referred object and generate an appropriate response.

Despite the rapid progress of Multimodal Large Language Models (MLLMs), such gesture-aware and region-specific question answering remains largely unexplored.
From examples in Fig.~\ref{fig:fig1_teaser}, we observe that even state-of-the-art MLLMs fail to address questions grounded in the user’s gesture.
This can be primarily attributed to two issues.
First, the datasets these MLLMs are trained on contain limited gesture-rich video data, especially those capturing natural pointing behavior and egocentric interactions between users and nearby objects. 
As a result, models rarely encounter examples where gestures and deictic language co-occur in realistic contexts.
Second, current architectures are not designed to explicitly encode or reason about gesture information. 
Most MLLMs integrate visual and textual inputs globally, without mechanisms to interpret hand positions or pointing directions. 
Consequently, they struggle to connect deictic expressions like “these” to the correct objects being referenced.

To address this, we introduce \datasetname, a dataset composed of 4,000 synthetic and 400 real egocentric videos.
These videos feature multiple objects and are specifically designed to support questions such as “What is the object I am pointing to?”, “How many objects of this type are there?”, or “What is the order of the objects I am pointing at?”.
To enable systematic evaluation, we benchmark several models across our deictic task categories.
%
Our evaluation benchmark contains 672 question–answer pairs over 300 real-world egocentric videos.
%
%
Unlike prior egocentric video question answering (VQA) datasets that emphasize general scene understanding, \datasetname focuses on fine-grained region-level reasoning, requiring both temporal and spatial comprehension to resolve gesture-based references accurately.

While training and benchmarking on this dataset helps mitigate the problem, the model must also explicitly attend to gestures.
To achieve this, we introduce \textbf{H}and \textbf{In}tent \textbf{T}okens (HINT), which are gesture-aware tokens, offering a compact representation and allowing the model to better interpret the user’s intent.
%
We obtain 3D hand keypoints for each frame of the video using an off-the-shelf hand reconstruction model, and pass them through a lightweight adapter to produce frame-aligned hand-intent tokens.

We then insert these tokens into the MLLM input sequence alongside the corresponding visual tokens, so the model receives an explicit gesture stream in addition to video and text. 
This design allows the MLLM to associate hand motion and pointing direction with the question and answer, leading to more precise grounding of deictic expressions (e.g., ``this one”).
In the experimental section, we demonstrate that training with gesture tokens on deictic questions improves performance by 6.5\% over standard fine-tuning, and surpasses the baseline by 8.6\%.

In summary, our contributions are the following:

\begin{itemize}
\item We \textbf{introduce} \datasetname, the first dataset specifically designed for deictic question answering in egocentric videos, where user gestures and pointing behaviors are central to interpretation. 

\item We \textbf{propose} \methodname, a simple yet effective approach that encodes gesture tokens derived from off-the-shelf 3D hand keypoints and interleaves them with the model input, providing explicit spatial and temporal context for interpreting pointing intent.

\item We \textbf{demonstrate} that \methodname achieves state-of-the-art performance on gesture-grounded question answering tasks, outperforming existing MLLMs and establishing a strong foundation for future research.
\end{itemize}

\begin{figure*}[t]
  \centering
  \includegraphics[width=0.9\linewidth]{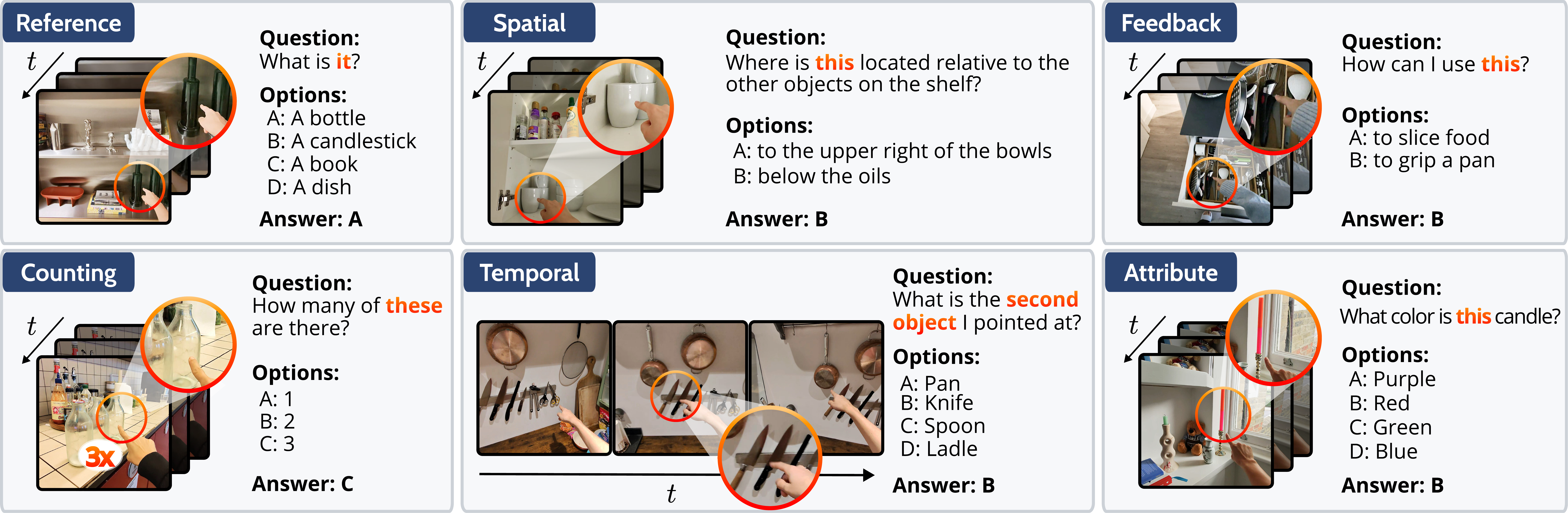}
   \caption{\textbf{Task taxonomy and examples from \datasetname.}
   \datasetname includes six subsets of questions regarding the properties of a pointed object.
    Each example shows egocentric video frames and a question using deictic references.
    Tasks include reference (object identification), counting (number of same objects), spatial (location and relative depth), temporal (order of multiple gestures), attribute (object properties), and feedback (object function).
    All questions require resolving deictic references through visual grounding of pointing gestures.
    The pointed objects are highlighted with red circles for visualization purposes.
   }
   \label{fig:task_taxonomy}
\end{figure*}

\section{Related Work}
\label{sec:rel_works}
\noindent{\textbf{Egocentric video question answering.}}
The field of egocentric video question answering (VQA) has emerged as a critical challenge, established by foundational large-scale egocentric video datasets~\cite{ego4d, epickitchens, hdepic,egoexo4d}.
Building upon these rich video corpora, dedicated VQA benchmarks were developed to specifically test complex reasoning about the wearer's actions and environment~\cite{qaego4d, qaego4dv2, egoschema, jia2022egotaskqa, easgbench,groundedquestionansweringlongegocentric,zhou2025egotextvqa,fan2019egovqa}.
Benchmarks such as EgoThink~\cite{egothink} and VidEgoThink~\cite{videgothink} focus on questions written in first-person perspective.
Recently, EgoGPT~\cite{egolifeegocentriclifeassistant} attempted to fine-tune MLLM~\cite{li2024llava} on wearer-perspective data, achieving improved performance on egocentric QA benchmarks~\cite{egolifeegocentriclifeassistant}.
Similarly, Ego-R1~\cite{egor1chainoftoolthoughtultralongegocentric} introduced a chain-of-tool-thought agent to reason over ultra-long first-person videos by decomposing queries and using external tools. These works mostly focus on long-term memory retrieval, habit analysis, or high-level queries.
However, none address region-specific ambiguities: current MLLMs fail to resolve deictic questions (e.g., ``what is this?") when the referent is only indicated via a pointing gesture. 

\noindent{\textbf{Region-specific visual question answering.}}
Region-level question answering has progressed rapidly in the image domain, with MLLMs like Ferret~\cite{ferret}, Osprey~\cite{osprey}, and DAM~\cite{lian2025damdescribe} grounded language to arbitrary image regions.
In video, methods like Artemis~\cite{artemis} track a specified region of interest across frames, while others like Elysium~\cite{elysiumexploringobjectlevelperception} and Omni-RGPT~\cite{omnirgptunifyingimagevideo} use key frames or learnable tokens to represent the region features.
Recently, large-scale datasets like VideoInfer~\cite{videorefersuiteadvancingspatialtemporal} have been built to train this capability.
However, all these works assume a referred region is explicitly given in the form of bounding box coordinates, segmentation masks, or a scribble.
In contrast, our work tackles the more natural scenario where the region of interest must be inferred implicitly from a human’s pointing gesture captured within the egocentric video itself.

\noindent{\textbf{Visual prompting for MLLMs.}} 
Providing additional visual cues or prompts has proven effective for steering MLLMs towards fine-grained understanding~\cite{visualpromptingmultimodallarge}.
These can be artificial overlays like alphanumeric tags~\cite{setofmarkpromptingunleashesextraordinary}, user-drawn scribbles ~\cite{vipllavamakinglargemultimodal, wang2025object}, or 2D points~\cite{pivotiterativevisualprompting}.
These methods prove that MLLMs can learn to interpret visual cues, but they all rely on artificial, non-natural prompts.
Our work departs from this paradigm by treating a natural human cue, already present in the egocentric video, through explicit hand tokens.
While other research has leveraged other natural signals like gaze to interpret the wearer's focus and align language with visual intent~\cite{yan2023voilaaaligningvisionlanguagemodels,egogazevqa}, our work tackles a distinct and complementary challenge of interpreting the pointing gesture.
Our approach of converting 3D hand keypoints into continuous hand tokens and interleaving them with the model's input sequence is a novel method for explicitly conditioning an MLLM on natural, pointing gestures for VQA.

\section{\datasetname Dataset}
We introduce \datasetname, the first dataset for pointing gesture-based question answering in egocentric video. 
The dataset combines both real-world and synthetic videos, each paired with multiple-choice question-answer pairs asking about specific regions and objects visible in the video.
We describe the task in \S\ref{sec:task}, the video collection in \S\ref{sec:collection}, and the question and answer generation in \S\ref{sec:qa_generation_pipeline}.

\subsection{Task Definition}
\label{sec:task}
Given an egocentric video containing one or more pointing gestures, where each gesture is associated with a target object at a specific timestamp, our task is to answer natural language deictic questions about the pointed objects.
A deictic question is formulated in first-person perspective and contains pronouns where the referent cannot be determined without visual and gestural context.
Common deictic expressions include demonstratives (e.g., ``this", ``that", ``those"), spatial indications (``here", ``there"), and temporal references (``the second object I pointed at").
Unlike standard video QA, where questions unambiguously describe the target object, deictic questions intentionally omit explicit descriptions, requiring the model to infer the referent from the user's pointing gesture.
To answer these questions, the system must jointly perform: (1) the spatial-temporal alignment between hand pose and objects to resolve what is being pointed at, (2) the linguistic grounding of deictic expressions to the visual scene, and (3) object properties, relations, and scene context to produce the correct answer.

\noindent{\textbf{Task taxonomy.}}
As illustrated in Fig.~\ref{fig:task_taxonomy}, we decompose deictic question answering into six question categories, each testing distinct reasoning capabilities.
\begin{itemize}
    \item \textit{Reference} – e.g., when asked “What is it?” while the user is pointing to an item on a shelf, the model must correctly identify this object.

    \item \textit{Counting} – determining the number of identical or similar objects in a view.

    \item \textit{Spatial} – understanding the relative position of the referenced object with respect to other objects in the scene.

    \item \textit{Temporal} – interpreting references when multiple pointing gestures occur in a sequence, using their order to resolve ambiguity.

    \item \textit{Feedback} – answering context-aware queries about an object’s function or relevance to the user’s goal.

    \item \textit{Attribute} – identifying properties such as color, shape, or material of the referenced object.
\end{itemize}
\vspace{0.5em}

\begin{figure}[t]
  \centering
  \includegraphics[width=\linewidth]{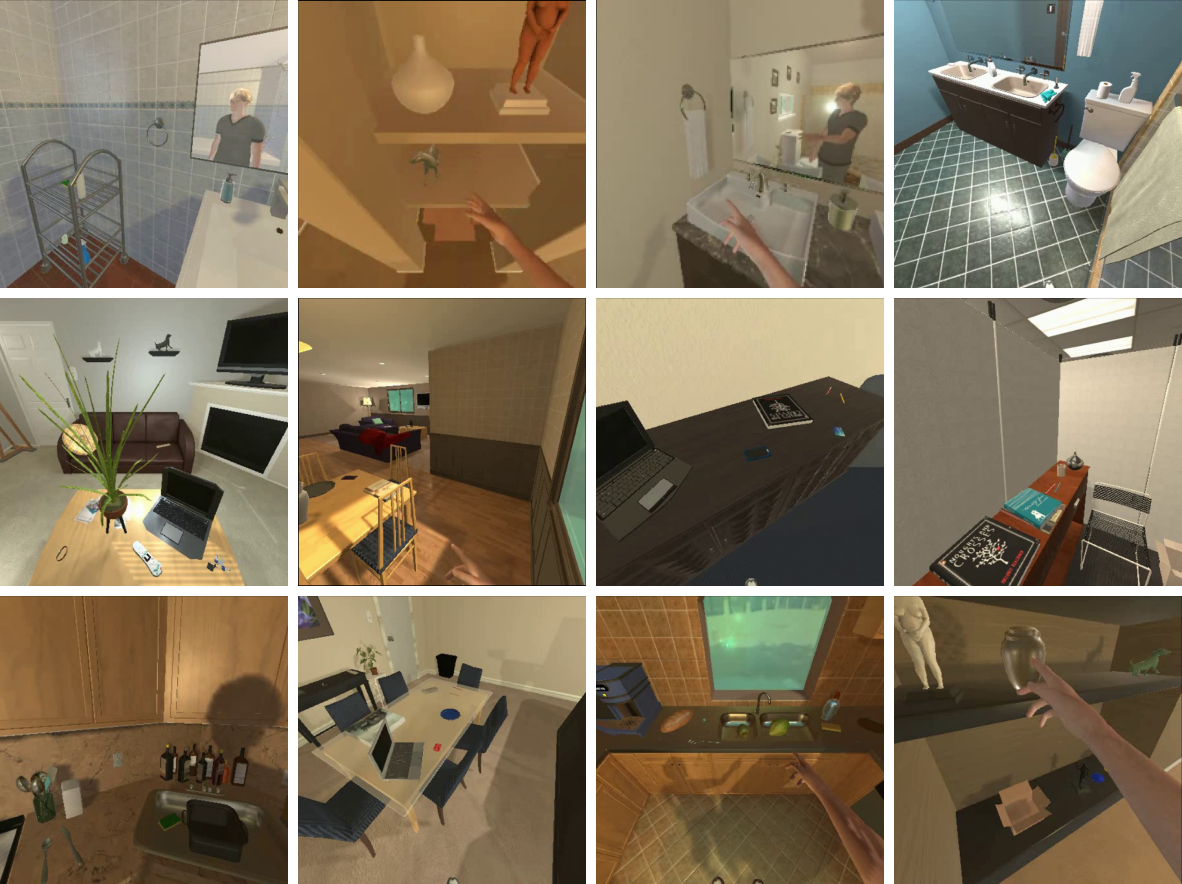}
   \caption{\textbf{Visualization of synthetic videos in \datasetname.} 
    Our synthetic data covers diverse indoor scenes with various lighting conditions.
   }
   \label{fig:exmaple_synthetic_videos}
\end{figure}

\subsection{Video Collection}
\label{sec:collection}
To construct a comprehensive dataset, we draw from both synthetic and real-world egocentric videos. Synthetic data allows us to overcome large-scale annotation challenges by providing precise control over object placement, viewpoint, and gesture timing. In contrast, real videos capture the natural variability and complexity of human behavior. Accordingly, our training set is predominantly synthetic with a small subset of real videos, while our test set comprises exclusively of real-world footage. 

\noindent{\textbf{Synthetic video generation pipeline.}}
We generate 4,000 synthetic videos using AI2-THOR \cite{ai2thor}, a photorealistic 3D simulator with 184 diverse indoor scenes.
From these scenes, we sample 12,000 viewpoints containing at least three visible nearby small objects, and select a subset of visible objects as target objects for pointing questions.
We create the pointing gestures by adapting MIXAMO animations~\cite{adobe_mixamo} using inverse kinematics to ensure the index finger aligns with the selected object. 
We then render the videos as 3–5 second clips at 30 frames per second (FPS) with $448 \times 448$ resolution.
Finally, we apply automatic quality filtering, retaining only those clips where the referenced object remains visible in at least 50\% of frames and the hand is reliably visible in over 60\% of the frames.
As visualized in Fig.~\ref{fig:exmaple_synthetic_videos}, generated videos show objects at diverse locations, with varied lighting.

\noindent{\textbf{Real video collection.}}
To construct a realistic evaluation scenario, we collect 400 egocentric videos using Meta Ray-Ban smart glasses~\cite{waisberg2024meta}.
We recruit 20 participants and instruct them to naturally point at objects in their daily environment, where more than 3 objects are visible.
These recordings took place across indoor (living room, kitchen, offices) and outdoor (streets, parks) settings (360 indoor and 40 outdoor).
Each clip ranges from 3–8 seconds with 30FPS, on average, at $1536\times2048$ resolution.
We allocate 100 videos for training (combined with synthetic data) and reserve 300 videos exclusively for evaluation.

\subsection{Question and Answer Generation}
\label{sec:qa_generation_pipeline}
\begin{figure}[t]
  \centering
  \includegraphics[width=\linewidth]{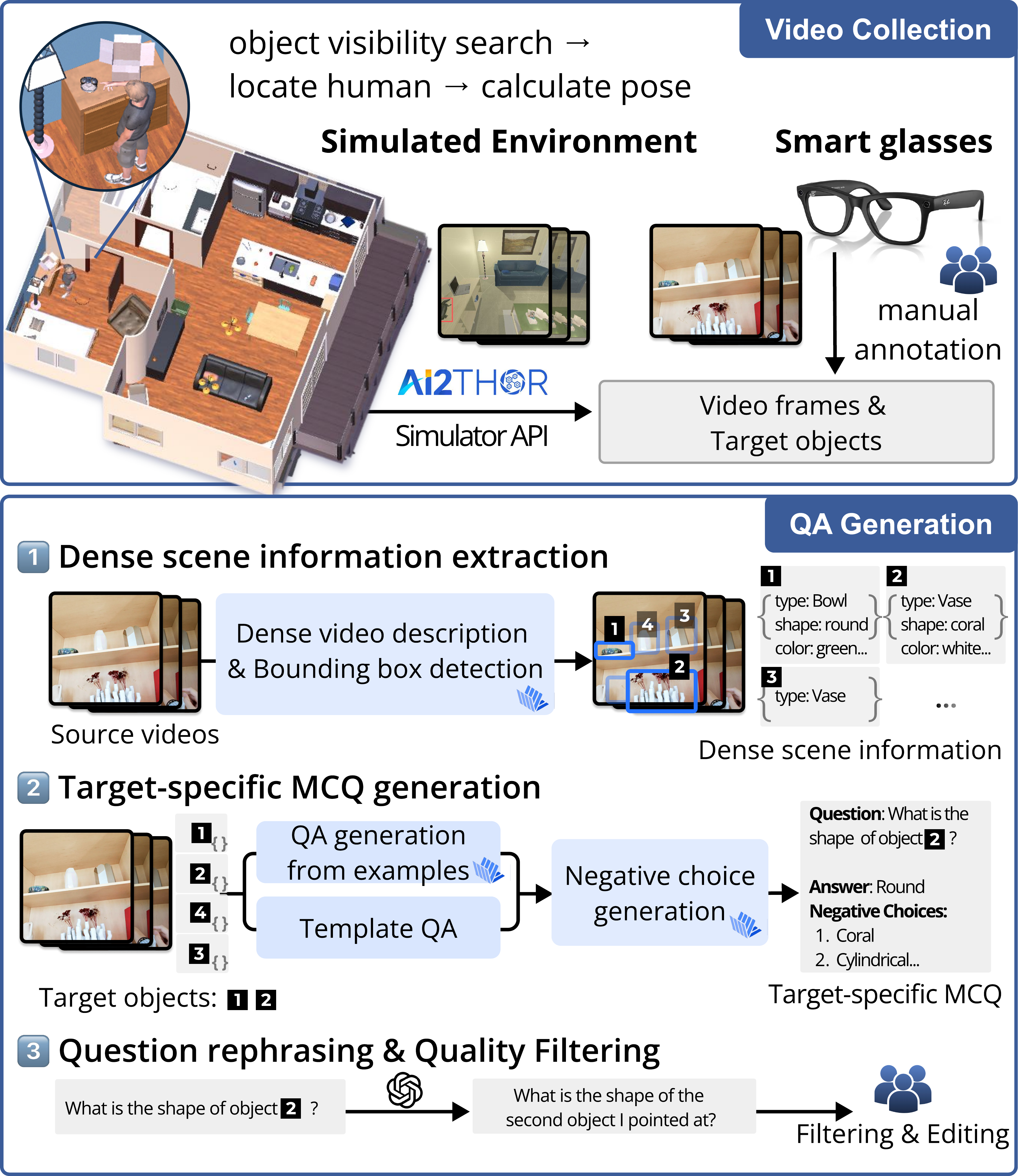}
   \caption{\textbf{\datasetname generation pipeline.}
   From a mixture of simulated and real egocentric videos, we automatically generate multiple-choice question answer pairs referring to the pointed objects in the video.
   The questions are made deictic so that the model should visually understand the pointing gesture to answer. 
   }
   \label{fig:datagen_pipeline}
\end{figure}

We design an automated pipeline to generate multiple-choice question-answer pairs from synthetic and real egocentric videos.
Generating descriptive questions about target objects requires two key capabilities: precise object localization across temporal sequences and a rich semantic understanding of spatial, temporal, and visual attributes.
To achieve this, our pipeline operates in three stages as illustrated in Fig.~\ref{fig:datagen_pipeline}: we first extract dense descriptions and metadata for the video, then we generate structured questions that reference the target objects, and finally, we transform the questions into natural first-person deictic expressions.
For real videos used in the evaluation, we perform manual verification and refinement to ensure high quality.

\noindent{\textbf{Stage 1: Extracting dense scene information.}}
The goal of this stage is to extract comprehensive scene information to automatically generate high-quality question-answer pairs.
The process differs between synthetic and real videos.
For synthetic videos, we leverage the simulator's API to extract depth maps, object-wise segmentation masks, categories, and 3D location.
Since the simulator does not provide all visual attributes (e.g., color), we supplement this by running an annotator MLLM (InternVL3-78B~\cite{zhu2025internvl3}) per object to extract these properties.
For real videos, we first generate scene metadata for all visible objects.
We follow a pipeline of SpatialRGPT~\cite{cheng2024spatialrgptgroundedspatialreasoning} to obtain dense object-wise scene descriptions with segmentation masks.
To establish a precise ground-truth for our pointing task, we manually annotate the bounding box of the specific target object (i.e., the object being pointed at).
We store this information as a list of JSON dictionaries, with each dictionary detailing a single object.

\noindent{\textbf{Stage 2: Target-specific multiple-choice question answer generation.}}
Using the comprehensive scene information from Stage 1, we generate multiple-choice question-answer pairs.
We first generate template-based question answer pairs that are specific to each of our six task categories.
Given example questions, videos, and the rule-based question-answer pairs, the annotator MLLM generates question-answer pairs that refer to the pointing target object using the object id placeholder.
For example, an `Attribute' question is generated using a template like `What color is \verb|<object2>|?'. We populate the negative answers based on the visible objects' metadata.
After constructing the structured question–answer pairs, we prompt the question-generating MLLM to produce a set of plausible hard negative options based on the scene information, visual input, and textually coherent alternatives.

\begin{figure}[t]
  \centering
  \includegraphics[width=\linewidth]{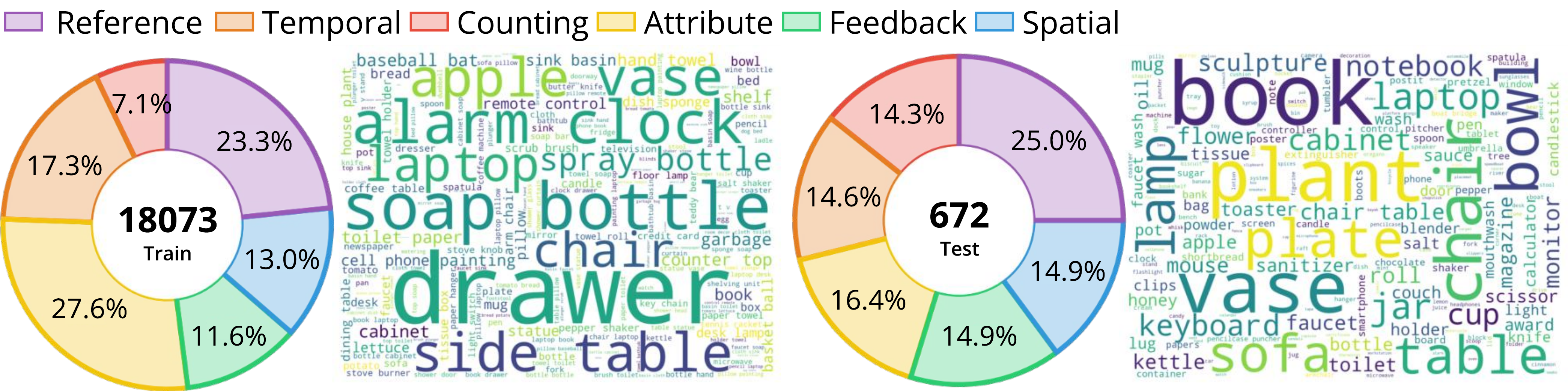}
   \caption{\textbf{\datasetname statistics.}
     Distribution of task types (charts) and common object word clouds for the (Left) training set (N=18073) and the (Right) test set (N=672).
   }
   \label{fig:data_statistics}
\end{figure}

\begin{figure*}[t]
  \centering
  \includegraphics[width=.9\linewidth]{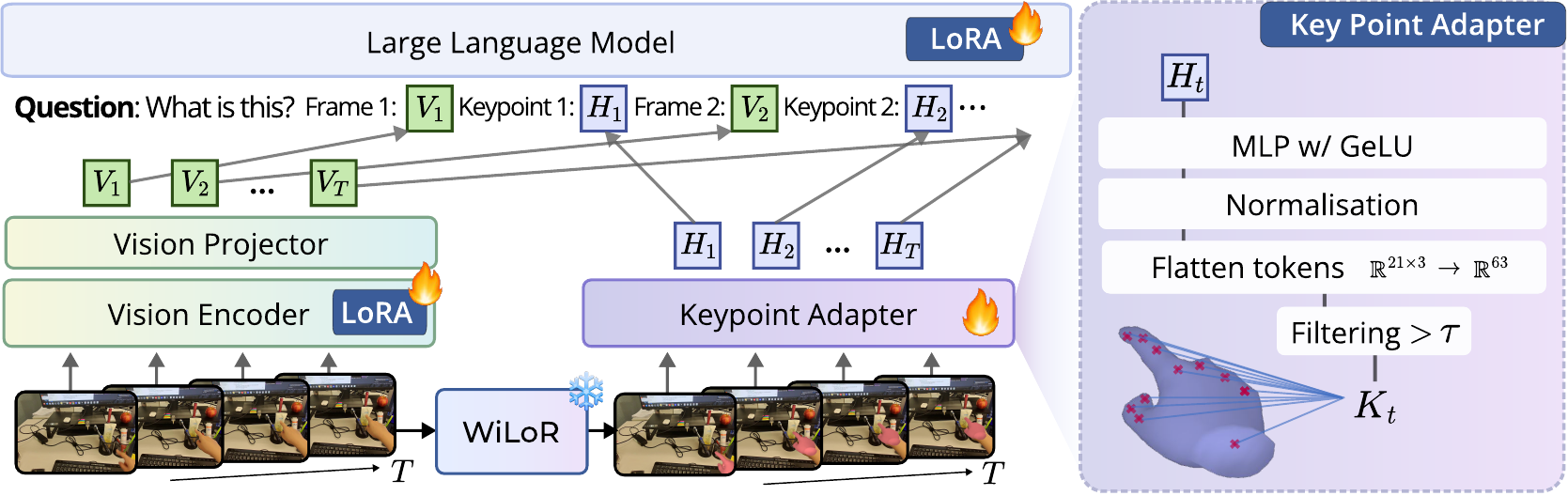}
   \caption{\textbf{\methodname overall architecture.} \methodname uses an additional adapter to model the 3D location and movement of the hand directly.
   $V_t$ denotes a visual token, $K_t$ the keypoint feature, and $H_t$ the hand intent token extracted for each frame $I_t$.
   }
   \label{fig:method}
\end{figure*}

\noindent{\textbf{Stage 3: Question rephrasing.}}
We convert the questions from Stage 2 into natural queries. We feed the multiple-choice QA pairs into GPT-4o~\cite{gpt4o}, which rephrases them by replacing object identifiers (e.g., \verb|<object2>|) with contextually appropriate deictic pronouns (e.g., ``this", ``it").

\noindent{\textbf{Quality control.}}
We manually inspect
every proposed question-answer pair for our 300 real-world evaluation videos.
We have two criteria: (1) correctness: the question and answer correspond with the video, (2) deictic ambiguity: the question is formulated using indirect pronouns. This ensures the question is difficult or impossible to answer correctly without understanding the pointing gesture.

\noindent{\textbf{Dataset statistics.}}
In Fig.~\ref{fig:data_statistics}, we visualize statistics and common object word clouds of \datasetname.
The dataset contains 4,000 synthetic videos and 400 real videos with a total of 18,745 question-answer pairs. It is split into two subsets: an instruction tuning (training) set that contains whole synthetic videos and their QA pairs, supplemented with 100 real videos and their 640 QA pairs, and the test set of 300 real videos with 672 QA pairs.

\section{\underline{H}and \underline{In}tent \underline{T}okens}
\label{sec:method}
Given an egocentric video, represented as a sequence of frames $\{I_1, I_2, ..., I_T\}$ and a deictic question in text, our goal is to generate the correct answer.
Current MLLMs struggle with such deictic queries because they often fail to (1) recognize that the question alone is ambiguous, and (2) accurately identify the user's referential intent behind the pointing gesture.
To address these challenges, we introduce \methodname, a model that processes the video in two parallel streams:  
a standard visual stream (\S\ref{sec:4_1_visual}) and a new hand-intent stream (\S\ref{sec:4_2_handtoken}).
As shown in Fig.~\ref{fig:method}, we develop a lightweight \emph{Keypoint Adapter} that converts 3D hand pose features from an off-the-shelf hand reconstruction model into a sequence of \emph{hand intent tokens}.
We then interleave these gesture tokens with the visual tokens (\S\ref{sec:4_3_interleaving}) and feed the combined sequence into the MLLM, providing explicit pointing information that helps resolve deictic ambiguity.
\subsection{Visual Token Extraction}
\label{sec:4_1_visual}
The primary visual stream follows the standard MLLM architecture. Each video frame $I_t$ is passed through a Vision Encoder (e.g., InternViT~\cite{zhu2025internvl3}) followed by a Vision Projector (an MLP).
This process maps the raw image into a sequence of embedding vectors (i.e., visual tokens), which we denote as $V_t$, and represents the visual content of the frame in the backbone LLM's embedding space.

\subsection{Hand Intent Token Extraction}
\label{sec:4_2_handtoken}
\noindent{\textbf{3D hand pose extraction.}}
Interpreting a referential gesture requires accurate estimation of the hand’s 3D pose.
This pose is the primary signal of referential intent, distinguishing a deliberate pointing action from other incremental hand motions. 
Consequently, our pipeline includes a 3D hand reconstruction module for estimating the hand pose in each frame. 
We choose to use WiLoR~\cite{potamias2025wilor} for this task since it has demonstrated robustness on in-the-wild images.
For each frame $I_t$, WiLoR outputs 3D camera space coordinates of 21 hand keypoints $K_t$. 
These keypoints provide a geometric representation of the hand's configuration per frame, which serves as the input to our Keypoint Adapter.

\noindent{\textbf{Keypoint adapter.}}
The role of this adapter is to project the 21 distinct 3D keypoints $K_t \in \mathbb{R}^{21 \times 3}$ into a single Hand Intent Token $H_t$ that holistically represents the entire posture of the hand for that frame:
%
%
\begin{align}
\tilde{k}_t &= \text{flatten}(K_t) \in \mathbb{R}^{63}, \\ H_t &=
\left\{
\begin{array}{ll}
    W_2\sigma(W_1 LN(\tilde{k}_t)), & \text{if } c_t \ge \tau \\
    \emptyset, & \text{otherwise,}
\end{array}
\right.
\label{eq:hand_intent_token}
\end{align}

\noindent where $W_1 \in \mathbb{R}^{d_h \times 63}$, $W_2 \in \mathbb{R}^{d \times d_h}$, $\sigma$ is GeLU, $c_t$ is the confidence of the hand detection, LN is LayerNorm, $d_h$ the hidden size, and $d$ matches the width of the LLM. 
Here, $\emptyset$ denotes the absence of a hand-intent token when $c_t < \tau$, i.e., no gesture token is inserted at time $t$.
This cheap adapter keeps latency low while exposing the LLM to compact frame-aligned gesture tokens.


\subsection{Frame-Keypoint Interleaving.}
\label{sec:4_3_interleaving}
We interleave hand intent tokens with visual tokens so the LLM can jointly reason over \emph{what} has happened and \emph{where the user has pointed}.
An example of a question and its answer in the dataset is as follows:

\textit{Question: What is this? A. a toothpaste B. a monitor ... Frame-1: $\langle vis \rangle$ Keypoint-1: $\langle key \rangle$ \dots\; Answer: A. }

For each $\langle vis \rangle$ token, we use the corresponding vision tokens $V_t$; likewise, for each $\langle key \rangle$ token, we use the corresponding keypoint tokens $H_t$ produced by the keypoint adapter. To cover the case where there is no hand present in the frame, we only interleave keypoint tokens if the detection confidence (given by WiLoR) exceeds $\tau=0.5$. This allows the model to naturally handle videos with intermittent hand visibility, a common occurrence in egocentric videos where hands move in and out of frame.

Since the HINT tokens are interleaved with the input sequence, the LLM will be temporally conditioned on them. For a sequence of length L, the probability of generating the target answers $\mathbf{X}_a$ is given by:
\begin{equation}
    p( X_{\texttt{a}} |  V, X_{\texttt{q}}, \textcolor{handtoken}{H}) =
    \prod_{i=1}^{L} p (x_i
| V, X_{\texttt{q}, <i}, X_{\texttt{a}, <i},
\textcolor{handtoken}{H_{<i}}),
    \label{eq:auto_regressive}
\end{equation}

\noindent where $X_{\texttt{q}, <i}$, $X_{\texttt{a}, <i}$, and $\textcolor{handtoken}{H_{<i}}$ denote the instruction, answer and \methodname tokens preceding the current prediction token. For the conditionals in~\eqref{eq:auto_regressive}, we explicitly add $\textcolor{handtoken}{H}$ to highlight that all answer tokens are grounded in the hand signal. This interleaved construction enables the LLM to jointly understand deictic context and the user’s temporally anchored references.

\begin{table*}[t]
\centering
    \begin{tabular}{lccccccccc}
    \toprule
    \textbf{Method} & \textbf{Size} & \textbf{LLM} & Refer. & Temporal & Spatial & Count & Attr. & Feed. & \textbf{Avg.} \\
    \midrule
    Random
           & - & - & 20.0 & 20.0 & 27.0 & 20.0 & 20.0 & 50.0 & 26.2 \\
    \midrule
    \textit{Proprietary Models} \\
    \textcolor{gray}{GPT-5~\cite{gpt5}}
           & \textcolor{gray}{-} & \textcolor{gray}{-} & \textcolor{gray}{75.6} & \textcolor{gray}{53.6} & \textcolor{gray}{62.3} & \textcolor{gray}{50.0} & \textcolor{gray}{56.1} & \textcolor{gray}{77.8} & \textcolor{gray}{62.6} \\

    \textcolor{gray}{GPT-4o~\cite{gpt4o}}
           & \textcolor{gray}{-} & \textcolor{gray}{-} & \textcolor{gray}{56.1} & \textcolor{gray}{29.5} & \textcolor{gray}{43.1} & \textcolor{gray}{44.8} & \textcolor{gray}{41.5} & \textcolor{gray}{65.7} & \textcolor{gray}{46.8} \\
    \midrule
    \textit{Open-source MLLMs $\geq$ 32B} \\
    \textcolor{gray}{Qwen3-VL~\cite{qwen3technicalreport}}
            & \textcolor{gray}{32B} & \textcolor{gray}{\textit{\small Qwen3}} & \textcolor{gray}{63.7} & \textcolor{gray}{67.9} & \textcolor{gray}{65.8} & \textcolor{gray}{66.7} & \textcolor{gray}{63.4} & \textcolor{gray}{77.2} & \textcolor{gray}{67.5} \\
\textcolor{gray}{InternVL2.5~\cite{internvl2d5}}
            & \textcolor{gray}{38B} & \textcolor{gray}{\textit{\small InternLM2.5}} & \textcolor{gray}{61.3} & \textcolor{gray}{57.1}  & \textcolor{gray}{60.5} & \textcolor{gray}{39.6} & \textcolor{gray}{63.4} & \textcolor{gray}{77.2} & \textcolor{gray}{59.9} \\

\textcolor{gray}{InternVL3~\cite{zhu2025internvl3}}
            & \textcolor{gray}{38B} & \textcolor{gray}{\textit{\small InternLM3}} & \textcolor{gray}{70.2} & \textcolor{gray}{67.9}  & \textcolor{gray}{65.8} & \textcolor{gray}{45.8} & \textcolor{gray}{65.9} & \textcolor{gray}{78.9} & \textcolor{gray}{65.8} \\
    \textcolor{gray}{LLaVA-OneVision~\cite{li2024llava}}              
            & \textcolor{gray}{72B} & \textcolor{gray}{\textit{\small Qwen2}} & \textcolor{gray}{61.3} & \textcolor{gray}{44.6} & \textcolor{gray}{60.5} & \textcolor{gray}{41.7} & \textcolor{gray}{51.2} & \textcolor{gray}{72.3} & \textcolor{gray}{55.3} \\
    \textcolor{gray}{InternVL3~\cite{zhu2025internvl3}}
            & \textcolor{gray}{78B} & \textcolor{gray}{\textit{\small InternLM3}} & \textcolor{gray}{71.4} & \textcolor{gray}{71.4} & \textcolor{gray}{62.3} & \textcolor{gray}{45.8} & \textcolor{gray}{68.3} & \textcolor{gray}{80.1} & \textcolor{gray}{66.6} \\
    \midrule
    \textit{Open-source MLLMs $\leq$ 8B} \\
    InternVL2.5~\cite{internvl2d5}
            & 8B & \textit{\small InternLM2.5} & 54.2 & 41.1  & 59.6 & 37.5 & 63.4 & 77.2 & 55.5 \\

    Qwen3-VL~\cite{qwen3technicalreport}
            & 8B & \textit{\small Qwen3} & 61.9 & 53.6  & 56.1 & 35.4 & 48.8 & 79.4 & 55.9 \\
    ViSpeak~\cite{fu2025vispeak}
            & 7B & \textit{\small Qwen2} & 65.5 & 42.9 & 48.2 & 39.6 & 51.2 & 70.2 & 52.9\\
    EgoGPT~\cite{egolifeegocentriclifeassistant}
            & 7B & \textit{\small Qwen2} & 67.3 & 46.4 & 50.9 & 47.9 & 48.8 & 74.1 & 55.9 \\
    VGLLM-QA~\cite{zheng2025learning}
            & 8B & \textit{\small Qwen2.5} & 57.7 & 35.7 & 53.5 & 39.6 & 36.6 & 70.2 & 48.9 \\
    \midrule
    LLaVA-OneVision~\cite{li2024llava}              
            & 7B & \textit{\small Qwen2} & 54.2 & 42.9 & 53.5 & 35.4 & 46.3 & 67.1 & 49.9 \\
    \rowcolor{cyan!15} \textbf{\methodname}$_\text{LLaVA-OneVision}$          
            & 7B & \textit{\small Qwen2} & \textbf{60.7} & \textbf{50.0}  & \textbf{56.1} & \textbf{39.6} & \textbf{48.8} & \textbf{71.1} & \textbf{54.4} \\
    InternVL3~\cite{zhu2025internvl3}
            & 8B & \textit{\small InternLM3} & 66.1 & 57.5 & 63.2 & 33.3 & 51.3 & 76.8 & 58.0 \\
    \rowcolor{cyan!15}
    \textbf{\methodname}$_\text{InternVL3-8B}$ & 8B & \textit{\small InternLM3} & \textbf{75.0} & \textbf{66.1}  & \textbf{64.9} & \textbf{35.4} & \textbf{61.0} & \textbf{79.8} & \textbf{63.7} \\ 
    \aboverulesepcolor{cyan!15}
    \midrule
    InternVL3~\cite{zhu2025internvl3}
            & 14B & \textit{\small InternLM3} & 63.1 & 66.1 & 61.4 & 50.0 & 58.5 & 77.2 & 62.7 \\
    \rowcolor{cyan!15}
    \textbf{\methodname}$_\text{InternVL3-14B}$
            & 14B & \textit{\small InternLM3} & \textbf{73.8} & \textbf{69.6} & \textbf{64.9}  & \textbf{54.2} & \textbf{63.4} & \textbf{82.5} & \textbf{68.1} \\
    \aboverulesepcolor{cyan!15}
    \bottomrule
    \end{tabular}
\vspace{-0.2cm}
\caption{\textbf{Performance of different MLLMs on the \datasetname test set.}
We report multiple-choice accuracy (\%). \methodname (highlighted in light blue) consistently improves its corresponding open-source backbones and outperforms all compared baselines. Task categories are Reference (Refer.), Temporal, Spatial, Counting (Count), Attribute (Attr.), and Feedback (Feed.). The random baseline reflects the varying number of answer choices across tasks.}
\label{tab:main_ourbench}
\end{table*}

\section{Experiments}
We conduct a series of experiments to establish the challenge of our \datasetname benchmark by evaluating several state-of-the-art MLLMs, and perform an analysis of our proposed \methodname method, ablating its core components.

\subsection{Experimental Setup}
\noindent{\textbf{Evaluation benchmark and metrics.}}
We conduct all experiments on the \datasetname test set, which consists of 300 real-world indoor and outdoor videos.
We report the multiple-choice accuracy (Acc. \%) for each of the six task categories along with the overall average accuracy.
For a fair comparison, all video MLLMs process 32 uniformly sampled frames per video.

\noindent{\textbf{Baseline models.}}
We evaluate 15 models spanning three categories: (1) proprietary models including GPT-5~\cite{gpt5}, and GPT-4o~\cite{gpt4o}; (2) Open-source generalist MLLMs, including Qwen3-VL~\cite{qwen3technicalreport} (8B, 32B), LLaVA-OneVision~\cite{li2024llava} (7B, 72B), InternVL2.5~\cite{internvl2d5} (8B, 38B), and InternVL3~\cite{zhu2025internvl3} (8B, 14B, 38B, 78B); and (3) models with relevant specializations including EgoGPT-7B~\cite{egolifeegocentriclifeassistant} (egocentric video understanding), VGLLM-QA-8B~\cite{zheng2025learning} (3D geometry understanding), and Vispeak~\cite{fu2025vispeak} (visual instruction).

\noindent{\textbf{\methodname implementation.}}
We implement \methodname on three representative backbones: LLaVA-OneVision-7B, InternVL3-8B, and InternVL3-14B.
We use WiLoR~\cite{potamias2025wilor} to extract 3D hand keypoints, setting 0.5 as the detection confidence threshold $\tau$.
We train the Keypoint Adapter from scratch, and fine-tune using LoRA~\cite{hu2021loralowrankadaptationlarge}  the vision encoder and LLM.
We optimize hyperparameters such as LoRA rank, alpha, and learning rate for each backbone individually.
We optimize all models using AdamW with a cosine schedule and a batch size of 32 for one epoch on our mixed synthetic-real training dataset. 
We provide a detailed list of the hyperparameters in the Appendix.

\subsection{Main Results}
\label{sec:5_main_results}
\noindent{\textbf{Zero-shot existing model performance.}}
The results in Table~\ref{tab:main_ourbench} show that \datasetname poses a challenge for any model. All models achieve an average accuracy below 70\%.
Among baseline models under 10B parameters, EgoGPT-7B achieves the highest reference accuracy at 67.3\%.
Particularly, comparing the Reference task with the Temporal tasks, even top models like GPT-5 show a dramatic drop. This shows that tracking multiple gestures across time requires fundamentally different capabilities than single gesture understanding.
We also observe that specialization provides limited benefits.
Despite strong performance on related benchmarks, models with relevant knowledge show small advantages in gesture understanding, e.g., EgoGPT reaches the best Reference and Counting accuracy among small models, but fails on temporal sequences.

\noindent{\textbf{Impact of model scaling.}}
Scaling within model families generally offers modest gain in performance.
For instance, the Reference accuracy of Qwen3-VL improves from 61.9\% (8B) to 63.7\% (32B), and InternVL3 improves from 66.1\% (8B) to 70.2\% (38B) to 71.4\% (78B).
This pattern suggests that while enhanced reasoning capability with a larger LLM backbone helps, the major bottleneck lies in the absence of explicit gesture understanding.

\noindent{\textbf{Performance improvement with \methodname.}}
As shown in Table \ref{tab:main_ourbench}, 
\methodname demonstrates consistent improvements across all backbones and task categories.
The method's primary impact is on the core grounding challenge, which is most evident in the Reference task. Here, we see the largest gains, such as +10.7 percentage points (\textit{pp}) for InternVL3-14B (63.1\% $\rightarrow$ 73.8\%) and +8.9 \textit{pp} for InternVL3-8B (66.1\% $\rightarrow$ 75.0\%).
This improved grounding also leads to gains in other tasks, though the effect varies by backbone. For instance, the InternVL3-8B also sees large boosts in the Temporal (+8.6\textit{pp}) and Attribute (+9.7\textit{pp}) tasks.
Similarly, \methodname improves LLaVA-OneVision, reaching an average improvement of 4.5\textit{pp} across the six tasks.

\noindent{\textbf{Computational cost and token overhead.}}
We measure inference time on 32-frame inputs using the InternVL3-8B backbone. Without hand keypoint extraction and the keypoint adapter, the inference takes 2.58s, while using the HINT tokens increases the time only to 2.84s. Moreover, with a threshold of $\tau = 0.5$, the HINT tokens account for less than 1\% of the total tokens fed to the LLM.

\begin{table}[]
\centering
\resizebox{1.0\linewidth}{!}{
\begin{tabular}{cccccc}
    \toprule
    \textbf{SFT} & \textbf{Hand Int.} & \textbf{Reference} & \textbf{Temporal} & \textbf{Spatial} & \textbf{Attribute} \\
    \midrule
    \xmark & \xmark & 66.1 & 57.5 & 63.2 & 51.3 \\
    \cmark & \xmark & 68.5 & 60.7 & 59.6 & 56.7 \\
    \rowcolor{cyan!15}
    \cmark & \cmark & \textbf{75.0} & \textbf{66.1} & \textbf{64.9} & \textbf{61.0} \\
    \aboverulesepcolor{cyan!15}
    \bottomrule
\end{tabular}
}
\caption{\textbf{Ablation of \methodname components.}  `SFT' denotes supervised fine-tuning on \datasetname. `Hand Int.' denotes use of our Hand Intent Token.
Combining both yields the largest gains.
}
\label{tab:ablation_components}
\end{table}

\subsection{Ablation Study and Analysis}
\label{sec:5_ablation_study}
For consistency, unless specified otherwise, we perform all ablations using InternVL3-8B.

\noindent{\textbf{Ablation of proposed components.}}
In Table~\ref{tab:ablation_components}, we analyze the contribution of each component of our method.
Supervised fine-tuning (SFT) on the \datasetname training data alone offers a small benefit, improving Reference accuracy from 66.1\% to 68.5\%.
However, combining SFT with  \methodname achieves 75.0\% on the same task.
This demonstrates that simply exposing the model to gesture-based questions is insufficient, thus the explicit architectural component to process the 3D hand pose signal is necessary.

\noindent{\textbf{Impact of training data composition.}}
In Table~\ref{tab:ablation_train_dataset}, we show how adding synthetic data to the real training set affects the performance.
Training on real videos only achieves 67.3\% on reference and 60.7\% on temporal tasks.
Mixing the two datasets yields the best results.

\begin{table}[]
\centering
\resizebox{1.0\linewidth}{!}{
\begin{tabular}{cccccc}
    \toprule
    \textbf{Real} & \textbf{Synthetic} & \textbf{Reference} & \textbf{Temporal} & \textbf{Spatial} & \textbf{Attribute} \\
    \midrule
    \xmark & \cmark & 69.0 & 62.5 & 60.5 & 58.5 \\
    \cmark & \xmark & 67.3 & 60.7 & 56.1 & 56.1 \\
    \rowcolor{cyan!15}
    \cmark & \cmark & \textbf{75.0} & \textbf{66.1} & \textbf{64.9} & \textbf{61.0} \\
    \aboverulesepcolor{cyan!15}
    \bottomrule
\end{tabular}
}
\vspace{-0.2cm}
\caption{\textbf{Impact of training dataset.}
We vary the usage of synthetic and real videos in the training set.
Using a synthetic dataset to complement a real dataset yields the best result.
}
\label{tab:ablation_train_dataset}
\end{table}
\noindent{\textbf{Hand intent modeling methods.}}
We test multiple methods to feed gesture information to the model, reported in Tab.~\ref{tab:ablation_viz_prompt_design}.
We test naive visual prompting strategies with two variations, where we plot the information of the user's hand onto the video frame, both of which are ineffective.
Plotting visual keypoints on the frames harms the performance on the reference task, while drawing visual arrows from the fingertip provides a moderate boost but is still suboptimal (70.2\%).
Input keypoint coordinates along with the video perform moderately (68.5-69.0\%  in the reference task).
Our learned keypoint encoder outperforms all alternatives (75.0\% reference), suggesting that allowing the model to learn how to process geometric hand information is more effective than providing it through visual or textual formats.

\begin{table}[]
\centering
\resizebox{1.0\linewidth}{!}{
\begin{tabular}{lccc}
    \toprule
    \textbf{Hand Intent Modeling} & \textbf{Reference} & \textbf{Temporal} & \textbf{Spatial} \\
    \midrule
    None & 68.5 & 60.7 & 59.6 \\
    \midrule
    Visual Keypoints & 57.1 & 60.7 & 61.4 \\
    Visual Arrow from Fingertip & 70.2 & 60.7 & 62.3 \\
    3D Keypoints in Text & 68.5 & 55.4 & 58.8 \\
    2D Keypoints in Text & 69.0 & 57.1 & 59.6 \\
    \midrule
    \belowrulesepcolor{cyan!15}
    \rowcolor{cyan!15} HINT & \textbf{75.0} & \textbf{66.1} & \textbf{64.9} \\
    \aboverulesepcolor{cyan!15}
    \bottomrule
\end{tabular}
}
\vspace{-0.2cm}
\caption{\textbf{Different methods of hand intent modeling.}
We compare different methods to encode the user's hand pose.
Our \methodname with learning keypoint adapter outperforms alternative representations, such as visual prompts and textual inputs.
}
\label{tab:ablation_viz_prompt_design}
\end{table}

\begin{table}[]
\centering
\resizebox{0.95\linewidth}{!}{
\begin{tabular}{cccccc}
    \toprule
    \textbf{$\tau$} & \textbf{Reference} & \textbf{Temporal} & \textbf{Spatial} & \textbf{Count} & \textbf{Attribute} \\
    \midrule
    0.1 & 66.7 & 58.7 & 62.9 & 32.4 & 59.3 \\
    0.3 & \textbf{75.0} & 58.7 & 62.9 & 35.3 & 59.3 \\
    \rowcolor{cyan!15}
    0.5 & \textbf{75.0} & \textbf{66.1} & \textbf{64.9} & \textbf{35.4} & \textbf{61.0} \\
    0.7 &  64.9 & 60.9 & 62.9 & 32.4 & 59.3 \\
    0.9 & 68.4 & 63.1 & 60.0 & 32.4 & 59.3 \\
    \bottomrule
\end{tabular}
}
\vspace{-0.1cm}
\caption{\textbf{Impact of hand detection confidence threshold $\tau$ on \methodname.}
 The threshold $\tau$ controls a trade-off between filtering noisy detections and retaining valid pointing gestures.
 A value of 0.5 achieves the best overall performance.
}
\label{tab:ablation_wilor_confidence}
\vspace{-0.1cm}
\end{table}

\begin{table}[]
\centering
\resizebox{1.0\linewidth}{!}{
\begin{tabular}{lcccc}
    \toprule
    \textbf{Method} & \textbf{Input Frames} & \textbf{Reference} & \textbf{Spatial} & \textbf{Attribute} \\
    \midrule
    InternVL3-8B & uniform 32 & 66.1 & 63.2 & 51.3 \\
    InternVL3-8B & keyframes & 61.3 & 62.3 & 61.0 \\
    \midrule
    InternVL3-14B & uniform 32 & 63.1 & 61.4 & 58.5 \\
    InternVL3-14B & keyframes & 60.1 & 56.1 & 61.2 \\
    \bottomrule
\end{tabular}
}
\vspace{-0.2cm}
\caption{\textbf{Impact of input frames.}
We compare uniform 32-frame input with keyframes manually selected as oracle, where each pointing gesture is most clearly visible (one frame per gesture). y
Even with this oracle advantage, Reference drops by 4.8/3.0\textit{pp} (8B/14B) compared to uniform 32-frame input.
}
\label{tab:ablation_keyframes}
\end{table}



\noindent{\textbf{Impact of hand detection confidence threshold $\tau$.}}
In Tab.~\ref{tab:ablation_wilor_confidence}, we analyze the impact of the hand detection confidence threshold $\tau$, which controls a trade-off between retaining valid gesture signals and filtering noise.
A low threshold ($\tau\leq$0.3) is too permissive, and the resulting noise harms performance on complex tasks like Temporal (58.7\%) and Attribute (59.3\%).
Conversely, a high threshold ($\tau\geq$0.7) is too strict, discarding valid gestures and causing Reference performance to drop (e.g., 64.9\% at $\tau$=0.7).
We find $\tau$=0.5 achieves the best overall balance, yielding the highest results across all categories.

\noindent{\textbf{Impact of input frames.}}
In Tab.~\ref{tab:ablation_keyframes}, we compare the model performance when using uniformly sampled frames versus keyframes.
For the keyframe condition, we feed the model the frames where the pointing gesture is most clearly visible (one keyframe per pointing gesture).
This setting tests whether a keyframe is sufficient to resolve deictic references.
The results demonstrate that, in practice, understanding pointing gestures requires temporal context beyond isolated keyframes, especially for the Reference task, where the model must directly identify the pointed object.
\section{Conclusion}
We introduced \datasetname, the first dataset tailored for gesture-grounded egocentric video question answering, featuring 4,000 synthetic and 400 real-world videos. 
The accompanying benchmark enables rigorous evaluation of fine-grained spatial and temporal reasoning from pointing gestures.
To advance gesture understanding, we proposed \methodnamefull(\methodname), a model that encodes 3D hand keypoints as tokens interleaved with visual and textual inputs, allowing effective interpretation of referential intent.
Our method achieves state-of-the-art performance, outperforming open-source baselines by an average of $5.4\%$. 
\section*{Acknowledgement}
S. Zafeiriou and R. Potamias was funded by the EPSRC Project GNOMON (EP/X011364/1) and Turing AI Fellowship (EP/Z534699/1). 
J. Deng was supported by the NVIDIA Academic Grant.

{
    \small
    \bibliographystyle{ieeenat_fullname}
    \bibliography{main}
}

\clearpage
\setcounter{page}{1}
\setcounter{section}{0}
\renewcommand{\thesection}{\Alph{section}}
\maketitlesupplementary

\section{Implementation Details}
\label{sec:supp_impl_detail}

\subsection{\methodname Training Details}
Table~\ref{tab:supp_impl_detail} summarizes the optimized hyperparameters used to finetune \methodname on each backbone.
We initialized the hyperparameter search around values commonly used in prior works and performed a grid search over LoRA rank and scaling factor pairs of \{(8, 16), (16, 32), (32, 64), (64, 128)\} and learning rates in \{2e-7, 1e-5, 2e-5\}.
The reported configurations were selected based on the best validation performance for each backbone.

\begin{table}[]
\centering
\resizebox{0.9\linewidth}{!}{
\begin{tabular}{lcccc}
    \toprule
    \textbf{Backbone} & \textbf{r} & \textbf{a} & \textbf{lr} & \textbf{Resolution} \\
    \midrule
    InternVL3-8B & 64 & 128 & 1e-5 & 448$\times$448 \\
    InternVL3-14B & 32 & 64 & 2e-5 & 448$\times$448 \\
    LLaVA-OneVision-7B & 32 & 64 & 1e-5 & 384$\times$384 \\ 
    \bottomrule
\end{tabular}
}
\caption{\textbf{Hyperparameters for \methodname per backbone.}
This table summarizes the optimized training configurations.
Here, `r' denotes the LoRA rank, `a' the LoRA scaling factor, and `lr' the learning rate used during finetuning.
}
\label{tab:supp_impl_detail}
\end{table}

In all experiments, we finetune only the LoRA adapters and the keypoint adapter while freezing the remaining backbone parameters.
We use AdamW as the optimizer with a cosine learning rate schedule and a linear warm-up ratio of 0.03. Input videos are uniformly sampled in time to a fixed number of 32 frames, resized to the resolution in Tab.~\ref{tab:supp_impl_detail}, and normalized following each backbone's default preprocessing.

\subsection{Evaluation Protocol}
\label{sec:supp_eval}
We evaluate all models using multiple-choice accuracy.
Each example consists of a question, candidate options, and the index of the correct option.
Models are instructed to answer by selecting an option letter (``A'', ``B'', etc.).
We handle several common output formats.
If the entire string (after trimming whitespace) is a single option letter, we take it as the prediction.
Otherwise, we progressively clean the output by stripping special markup such as \texttt{\textless answer\textgreater ...\textless/answer\textgreater}, as well as trailing control tokens (e.g., \texttt{\textless|im\_end|\textgreater}) and boilerplate phrases (e.g., ``Here's the answer:'').
After this cleaning step, we apply a sequence of regular expressions that search for:
(i) a letter enclosed in parentheses, e.g., ``(A)'';
(ii) a letter followed by punctuation, e.g., ``A.'', ``B)'' or ``C]''; and
(iii) an ``Answer:'' pattern at the end of the string, e.g., ``Answer: D''.
In all cases, we restrict matches to letters between ``A" and the last valid option letter for that specific example.
If a match is found and the letter is valid, we treat it as the model's prediction and otherwise, we mark the prediction as invalid.
Multiple-choice accuracy is computed as the fraction of examples for which the extracted option letter exactly matches the ground-truth option.
For all baselines and \methodname variants reported in the paper, we successfully extracted a valid option letter for every prediction, so the reported accuracies correspond to exact letter-wise matches without any manual corrections or post-hoc filtering.

\section{Additional Analysis}
\begin{table}[t]
\centering
\resizebox{1.0\linewidth}{!}{
\begin{tabular}{lcccc}
    \toprule
    \textbf{Method} & Video-MME & MVBench & EgoSchema & EgoBlind \\
    \midrule
    InternVL3-8B & 64.2 & 73.2 & 67.2 & 52.8 \\
    \methodname$_\text{InternVL3-8B}$ & 64.6 & 73.2 & 67.1 & 57.5 \\
    \bottomrule
\end{tabular}
}
\caption{\textbf{Performance on existing video understanding benchmarks.}
Comparison of \methodname against the baseline InternVL3-8B on Video-MME~\cite{videomme}, MVBench~\cite{mvbench}, EgoSchema~\cite{egoschema} (MCQ), and EgoBlind~\cite{xiaoegoblind}.
The results indicate that finetuning on our dataset preserves the model's general video understanding capabilities.
}
\label{tab:supp_existing_benchmarks}
\end{table}

\subsection{\methodname Performance on Existing Benchmarks}
\label{sec:supp_existing_benchmarks}
In Table~\ref{tab:supp_existing_benchmarks}, we report the performance of \methodname on standard video understanding benchmarks to investigate whether finetuning on our proposed dataset compromises the model's pretrained general capabilities.
We compare our method with the backbone model, InternVL3-8B, on Video-MME~\cite{videomme}, MVBench~\cite{mvbench}, and EgoSchema~\cite{egoschema}.
All evaluations use 32 uniformly sampled frames without any additional fine-tuning on the target benchmarks.
This result demonstrates that the gesture-aware representations learned by \methodname transfer positively to related egocentric assistive QA scenarios, even when explicit pointing gestures are absent in the target benchmark.
As can be easily observed, \methodname achieves comparable performance compared to the baseline.
This demonstrates that our finetuning strategy effectively injects pointing gesture understanding capability without leading to catastrophic forgetting in general video understanding tasks, and also can be transfer positively to related egocentric assistive QA scenarios even when explicit pointing gestures are absent in the target benchmark.

\subsection{Human Performance}
\label{sec:supp_human_performance_rebuttal}
We report the average accuracy of 5 participants.
Each participant evaluated the full test set.
The near-ceiling performance (total average 95.9\%) confirms the questions are clear and easy for humans, yet reveals a substantial gap to current MLLMs (best performing model 68.1\%).
\begin{table}[t]
\centering
\resizebox{\columnwidth}{!}{%
\begin{tabular}{rcccccc}
\hline
\multicolumn{1}{r}{}
& \multicolumn{1}{c}{\textbf{Ref.}}
& \multicolumn{1}{c}{\textbf{Temp.}}
& \multicolumn{1}{c}{\textbf{Spat.}}
& \multicolumn{1}{c}{\textbf{Count}}
& \multicolumn{1}{c}{\textbf{Attr.}}
& \multicolumn{1}{c}{\textbf{Feed.}} \\ 
\cline{1-7}
    Human & 98.2 & 94.3 & 93.3 & 92.8 & 96.5 & 100.0 \\
\hline
\end{tabular}%
}
\vspace{-1em}
\caption{\textbf{Human performance on \datasetname.}
We report the average accuracy of 5 human participants, each evaluating the full test set (672 questions).
}
\label{tab:rebuttal_humaneval}
\end{table}

\subsection{Effect of Hand Gestures in Video}
\begin{table}[t]
\centering
\resizebox{\linewidth}{!}{
\begin{tabular}{lcccc}
    \toprule
    \textbf{Video} & Reference & Temporal & Spatial & Attribute \\
    \midrule
    Original & 75.0 & 66.1 & 64.9 & 61.0 \\
    w/o Hand & 41.7 & 21.4 & 44.4 & 36.3 \\
    \bottomrule
\end{tabular}
}
\vspace{-1em}
\caption{\textbf{Ablation study on the effect of hand gestures.}
Performance comparison on our dataset with and without the pointing hand visible in the video frames.
The significant drop in performance across all tasks in the `w/o Hand' setting confirms that the pointing gesture is essential for identifying the target.
}
\label{tab:supp_wo_hand}
\end{table}
To assess the importance of explicit pointing cues, we construct a static-video variant of \datasetname in which the camera wearer's hand is removed while keeping the rest of the scene and camera motion unchanged.
We then evaluate models, including \methodname, on this modified data using the same training and evaluation protocol as in the main experiments.
Table~\ref{tab:supp_wo_hand} shows that removing hand gestures causes a large degradation in performance.
This confirms that egocentric pointing cues are critical for resolving deictic references in \datasetname.

\subsection{Dataset Bias Analysis}
\label{sec:supp_dataset_bias}

\begin{table}[t]
\centering
\resizebox{\linewidth}{!}{
\begin{tabular}{lcccccc}
    \toprule
     & Reference & Temporal & Spatial & Count & Attribute & Feedback \\
    \midrule
    Random       & 20.0 & 20.0 & 27.0 & 20.0 & 20.0 & 50.0 \\
    Blind        & 16.7 & 25.5 & 32.0 &  6.3 & 22.5 & 54.9 \\
    Choices-only & 20.8 & 25.5 & 20.8 & 10.4 & 21.6 & 54.1 \\
    \bottomrule
\end{tabular}
}
\caption{\textbf{Dataset bias analysis.}
We evaluate two text-only baselines to detect potential shortcuts: ``Blind'' receives the question text without video, and ``Choices-only'' receives only the answer options without the question or video.
Performance near random chance confirms that \datasetname requires visual grounding.
}
\label{tab:supp_dataset_bias}
\end{table}

To verify that \datasetname does not contain unintended textual shortcuts, we evaluate two degenerate baselines that receive no visual input (Table~\ref{tab:supp_dataset_bias}).
The \textbf{Blind} baseline feeds only the question text (without the video) to the model, testing whether the question wording alone leaks the answer.
The \textbf{Choices-only} baseline provides only the multiple-choice options (without either the question or video), testing whether the answer can be guessed from the option distribution.
Both baselines perform near random chance across all six task categories: Blind achieves 16.7--54.9\% and Choices-only achieves 10.4--54.1\%, closely matching the random baseline of 20.0--50.0\%.
These results confirm that \datasetname genuinely requires visual grounding of pointing gestures to resolve deictic references, and that our question-answer generation pipeline does not introduce systematic textual or statistical shortcuts.

\subsection{Failure Analysis}
\label{sec:supp_failure_analysis_rebuttal}
\begin{figure}[t]
  \centering
  \includegraphics[width=\linewidth]{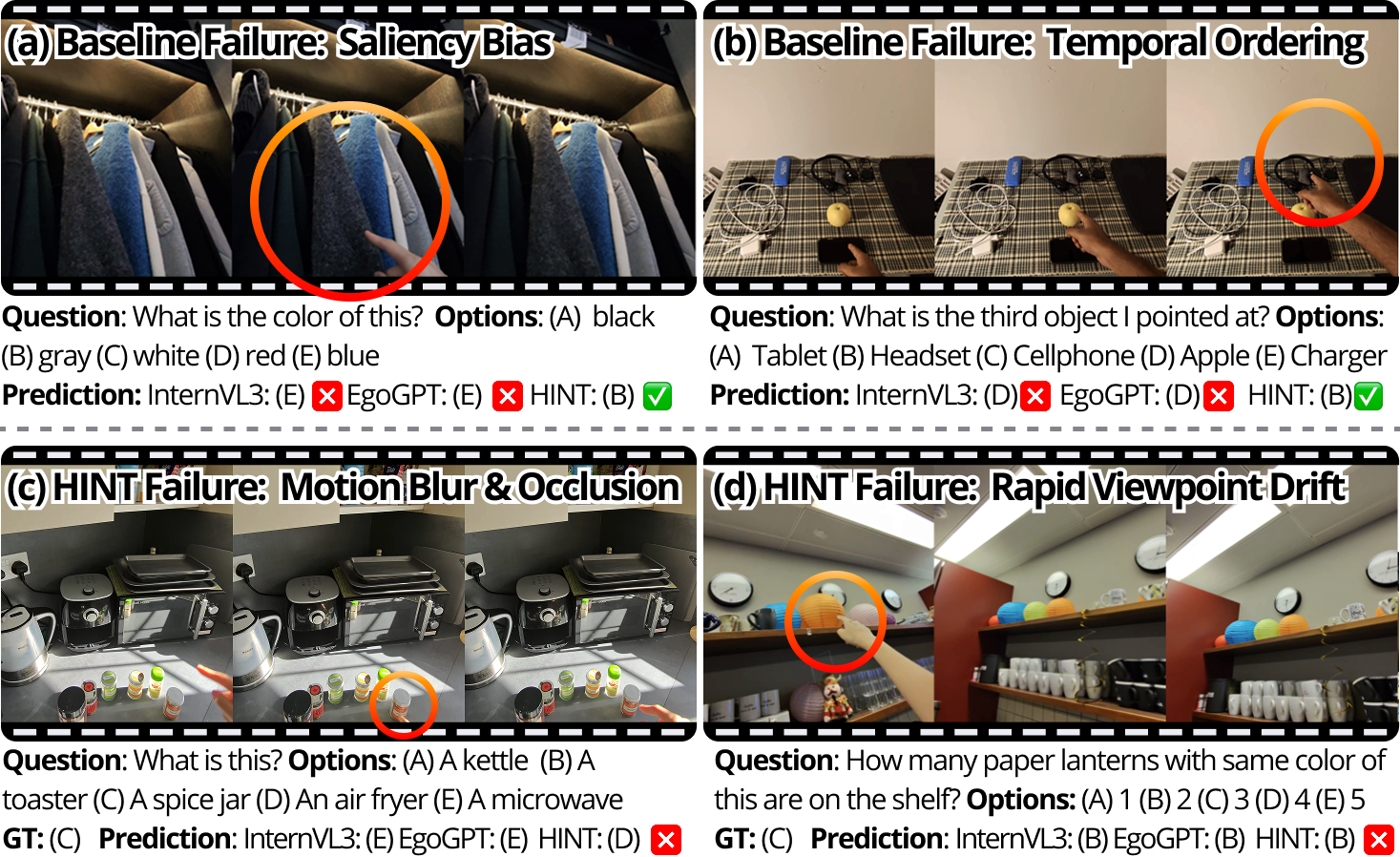}
  \caption{\textbf{Representative failure cases on \datasetname.}
    (a)-(b): baseline MLLM failures due to saliency bias and temporal confusion, respectively.
  (c)-(d): remaining \methodname failures caused by unreliable hand keypoints and rapid viewpoint drift.
  }
  \label{fig:supp_failure_analysis}
\end{figure}

In Figure~\ref{fig:supp_failure_analysis}, we visualize example failure modes to better understand the limitations of both existing MLLMs and our proposed \methodname.

\noindent\textbf{Existing MLLM failure modes.}
We identify two dominant error patterns in baseline models.
(a)~\textit{Saliency/center bias}: in cluttered scenes, models tend to predict a visually prominent or centrally located object rather than the one actually being pointed at.
For example, the model may focus on a nearby blue shirt instead of the actual referent.
(b)~\textit{Temporal confusion}: when multiple objects are sequentially pointed at, baseline models frequently confuse the temporal order, e.g., predicting the second pointed object when the question asks about the first.
\methodname mitigates both failure modes by providing frame-aligned 3D hand geometry that explicitly anchors deictic references to the correct spatial and temporal locations.

\noindent\textbf{Remaining \methodname failure modes.}
Despite the overall improvements, \methodname fails under certain challenging conditions.
(c)~\textit{Unreliable gesture signal}: when the hand is affected by motion blur or partial occlusion, the 3D hand reconstruction from WiLoR becomes noisy.
In such cases, the hand detection confidence may fall below the threshold $\tau$, resulting in absent hand intent tokens, or the tokens may encode misleading pose information.
(d)~\textit{Rapid viewpoint drift}: fast head movements can cause the target object to leave the camera's field of view entirely, making it harder for the model to associate the gesture with the correct referent regardless of the quality of hand tokens.
These remaining errors are largely attributable to input signal quality rather than architectural limitations, suggesting that advances in robust hand pose estimation and temporal object tracking under egocentric motion would yield further gains.

\begin{table}[b]
\centering
\resizebox{1.0\linewidth}{!}{
\begin{tabular}{lcccc}
    \toprule
    Data Split & \# Vid. & \# QA & Vid. Dur.(s) & \# Obj. \\
    \midrule
    Train(Real) & 100 & 640 & 4.61 & 22.5 \\
    Train(Synthetic) & 4,000 & 18,745 & 11.6 & 11.6 \\
    Test (Real) & 300 & 672 & 5.05 & 16.5 \\
    \bottomrule
\end{tabular}
}
\caption{\textbf{\datasetname statistics.} 
Statistics of the dataset across synthetic and real-world splits.
Real-world clips generally feature higher object density (scene complexity) than synthetic clips.}
\label{tab:supp_data_statistics}
\end{table}

\begin{figure}[b]
  \centering
  \includegraphics[width=0.7\linewidth]{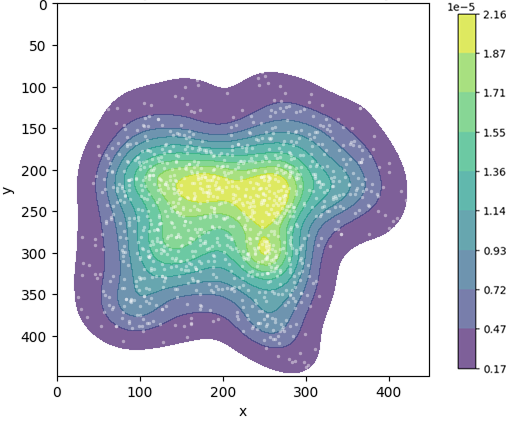}
   \vspace{-1em}
   \caption{\textbf{Spatial distribution of target objects in \datasetname.} `x' and `y' axes refer to the horizontal and vertical pixel coordinates, respectively.
   Individual object locations are shown as scattered dots.
   }
   \label{fig:supp_obj_loc_heatmap}
\end{figure}

\section{\datasetname}
\label{sec:supp_dataset_viz}
In this section, we provide additional analysis of \datasetname, qualitative visualizations, and a detailed description of the question answer pair generation pipeline.

\subsection{Video Collection Details}
\label{sec:supp_video_collection}

\noindent\textbf{Participants.}
We recruit 20 participants from 12 nationalities (11 female, 9 male; ages 18--45) to collect the real-world portion of \datasetname.
The diverse participant pool ensures variation in hand shape, skin tone, and pointing style, which encourages models to generalize across different users rather than overfitting to a narrow demographic.

\noindent\textbf{Scenes and activities.}
Real-world videos are captured across a broad variety of settings, including offices, kitchens, living rooms, streets, train platforms, and balconies.
Activities during recording span desk work, organizing, artwork, and cooking, reflecting natural everyday scenarios in which pointing gestures commonly occur.
Of the 400 collected videos, 360 are recorded in indoor environments and 40 in outdoor settings.

\noindent\textbf{Capture protocol.}
Each participant wears Meta Ray-Ban smart glasses and is instructed to point with one hand using an extended index finger for at least 2 seconds at objects of their choice.
We require that more than 3 objects are visible in the scene to ensure sufficient visual complexity for generating challenging deictic questions.
Each clip ranges from 3--8 seconds at 30 FPS with a resolution of 1536$\times$2048.

\subsection{Dataset Statistics}
Table~\ref{tab:supp_data_statistics} summarizes the scale and complexity of \datasetname. 
The dataset is composed of a large-scale synthetic training set (4,000 videos) to encourage robust generalization, complemented by real-world data for domain adaptation and testing. 
Notably, the real-world videos contain a higher density of objects (avg. 22.5 and 16.5 per video) compared to the synthetic set (avg. 11.6 per video), presenting a greater challenge in grounding target pointed objects.

In Fig.~\ref{fig:supp_obj_loc_heatmap}, we visualize the spatial distribution of target objects by their center coordinate within the frame. 
While there is a natural center bias typical of egocentric videos, where users tend to center objects they interact with, the heatmap demonstrates a diverse distribution of the object location. 
This broad distribution confirms that models cannot rely solely on center priors and must utilize the pointing gestures to resolve references.


\subsection{Extended Visualization of \datasetname}
We provide additional qualitative examples from the real-world split of \datasetname in Figures~\ref{fig:supp_real_video_frames_indoor} and \ref{fig:supp_real_video_frames_outdoor}.
Figure~\ref{fig:supp_real_video_frames_indoor} presents samples recorded in \textbf{indoor environments}, illustrating various household objects and close-range pointing gestures.
In contrast, Figure~\ref{fig:supp_real_video_frames_outdoor} depicts sequences captured in \textbf{outdoor settings}. 
These visualizations showcase the range of scenes and contexts included in the collected data.

\subsection{Synthetic Video Generation}






We generate synthetic egocentric videos using the AI2-THOR simulator~\cite{ai2thor} running in Unity.
Our implementation utilizes AI2-THOR commit \texttt{0+8524ea} and Unity version \texttt{2020.3.25f1}
The video generation process follows four steps.

\noindent First, we randomly sample a scene, an agent location, and a head camera orientation.
We keep only viewpoints where at least three objects are simultaneously visible according to ground-truth instance segmentation masks from the simulator.

\noindent From the visible entities, we select a target object of the pointing gesture.
To ensure valid grounding, we require the object to meet two criteria: more than 10\% of its pixels must be visible, and its depth must be within 2.0 meters from the agent.

\noindent For the target object, we retrieve a pointing animation from the MIXAMO~\cite{adobe_mixamo} library that possesses the most similar initial pose to the required trajectory.
We then apply Inverse Kinematics (IK) to refine the avatar's arm and hand configuration, forcing the index finger to align precisely with the target object.

\noindent After IK adjustment, we verify that the pointing ray originating at the fingertip intersects the 3D bounding box of the target. If the pointing deviates or misses the target, we discard this sample and resample a new viewpoint/pose.

\subsection{Automatic Question and Answer Generation Details}
\label{supp:qagen_prompts}
We provide a detailed breakdown of the prompts and logic used in our three-stage generation pipeline described in the main paper (\S~\refmain{3.3}). We utilize InternVL3-78B~\cite{zhu2025internvl3} for scene information extraction and multiple-choice question-answer pair generation (Stages 1\& 2) and GPT-4o~\cite{gpt4o} for linguistic refinement and quality control (Stage 3).

\begin{figure*}
    \centering
    \includegraphics[width=\linewidth]{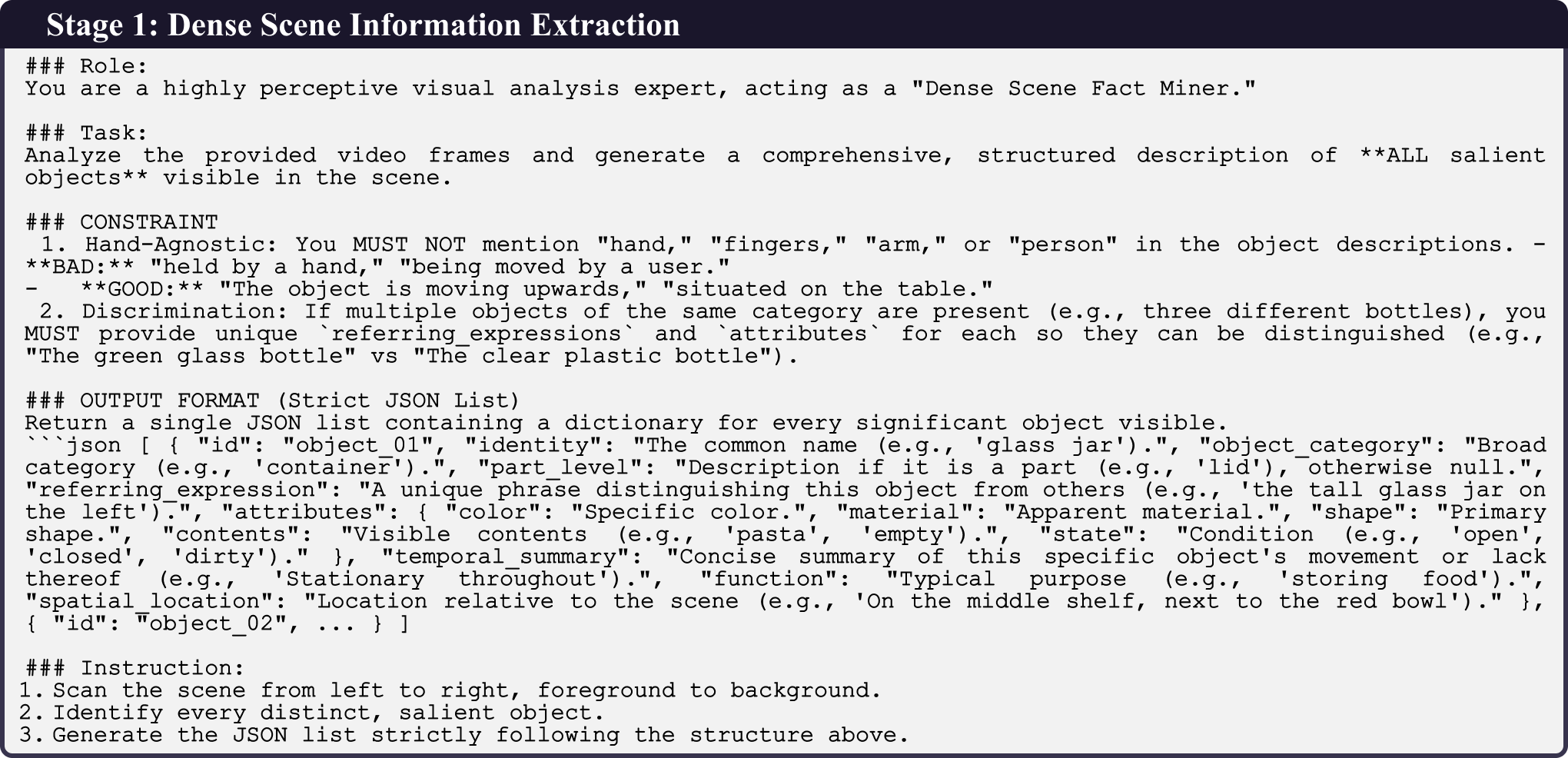}
    \vspace{-2em}
    \caption{\textbf{Prompt used for dense scene information extraction (Stage 1).}
    Instructions given to the MLLM to extract a structured JSON list of all salient objects.}
    \label{fig:supp_prompt_stage1_info}
\end{figure*}

\noindent{\textbf{Stage 1: Dense scene information extraction.}} Instead of generic captions, we require structured scene graphs to ground the subsequent QA generation. As shown in Fig.~\ref{fig:supp_prompt_stage1_info}, we prompt the annotator MLLM to act as a Dense Scene Fact Miner.'' The model analyzes the video frames and outputs a JSON list of all salient objects. To ensure high-quality grounding, we enforce two strict constraints in the prompt:
(1) Hand-Agnostic Descriptions: The model is explicitly forbidden from describing human interaction (e.g., pointed by a hand) to ensure object attributes are described objectively based on their visual state. (2) Discriminative Expressions: If multiple instances of the same category exist, the model must generate unique referring expressions to distinguish them.

\noindent{\textbf{Stage 2: Target-specific multiple-choice question answer generation.}} 
We generate the question stem and the multiple-choice options in two sequential sub-steps to ensure logical consistency.

\begin{figure*}
    \centering
    \includegraphics[width=\linewidth]{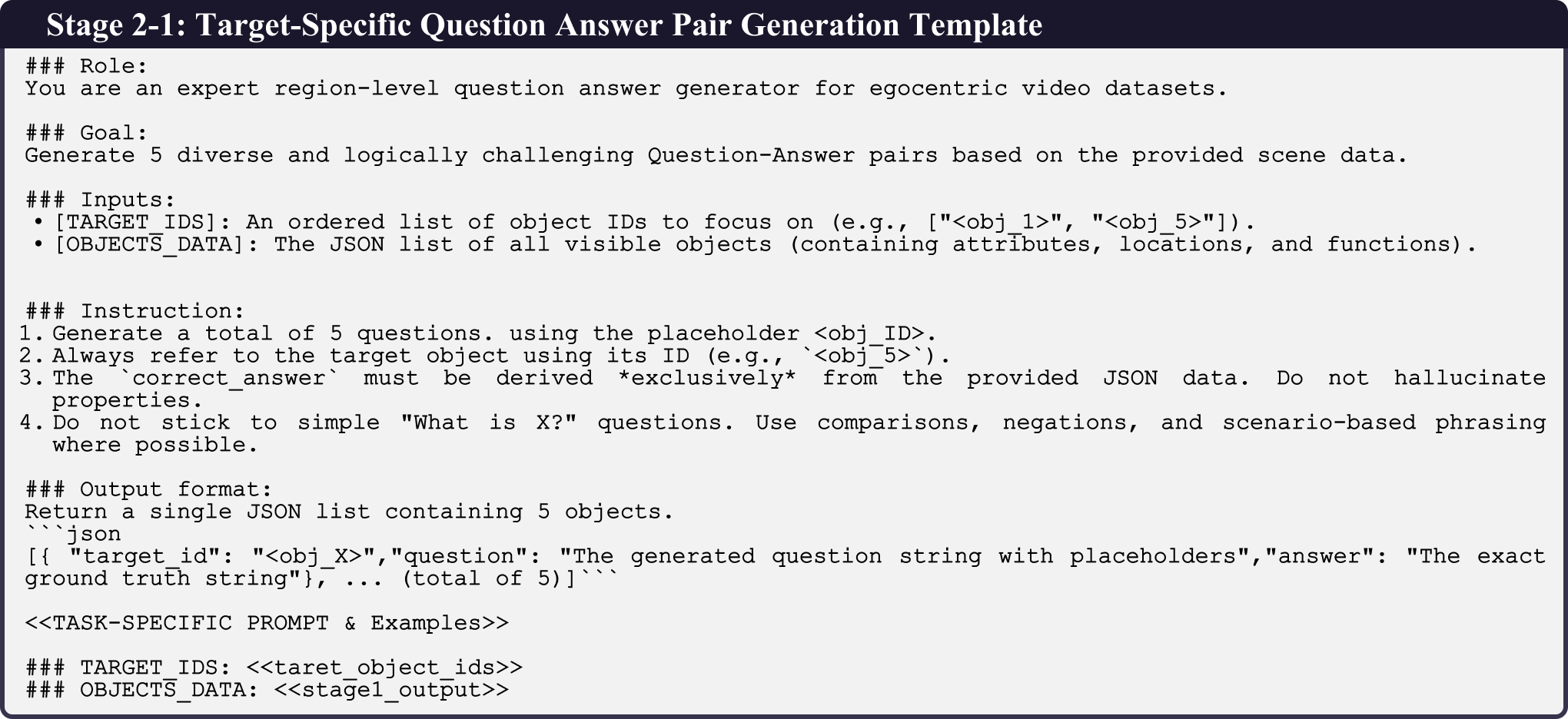}
    \vspace{-2em}
    \caption{\textbf{General template for QA pair generation (Stage 2-1).}
    The base instruction template is used for all question types. It enforces that the "correct answer" must be derived exclusively from the scene JSON provided in Stage 1 to prevent hallucinations.}
    \label{fig:supp_prompt_stage2_1_general}
\end{figure*}

\begin{figure*}[t]
    \centering
    \vspace{-2em}
    \includegraphics[width=\linewidth]{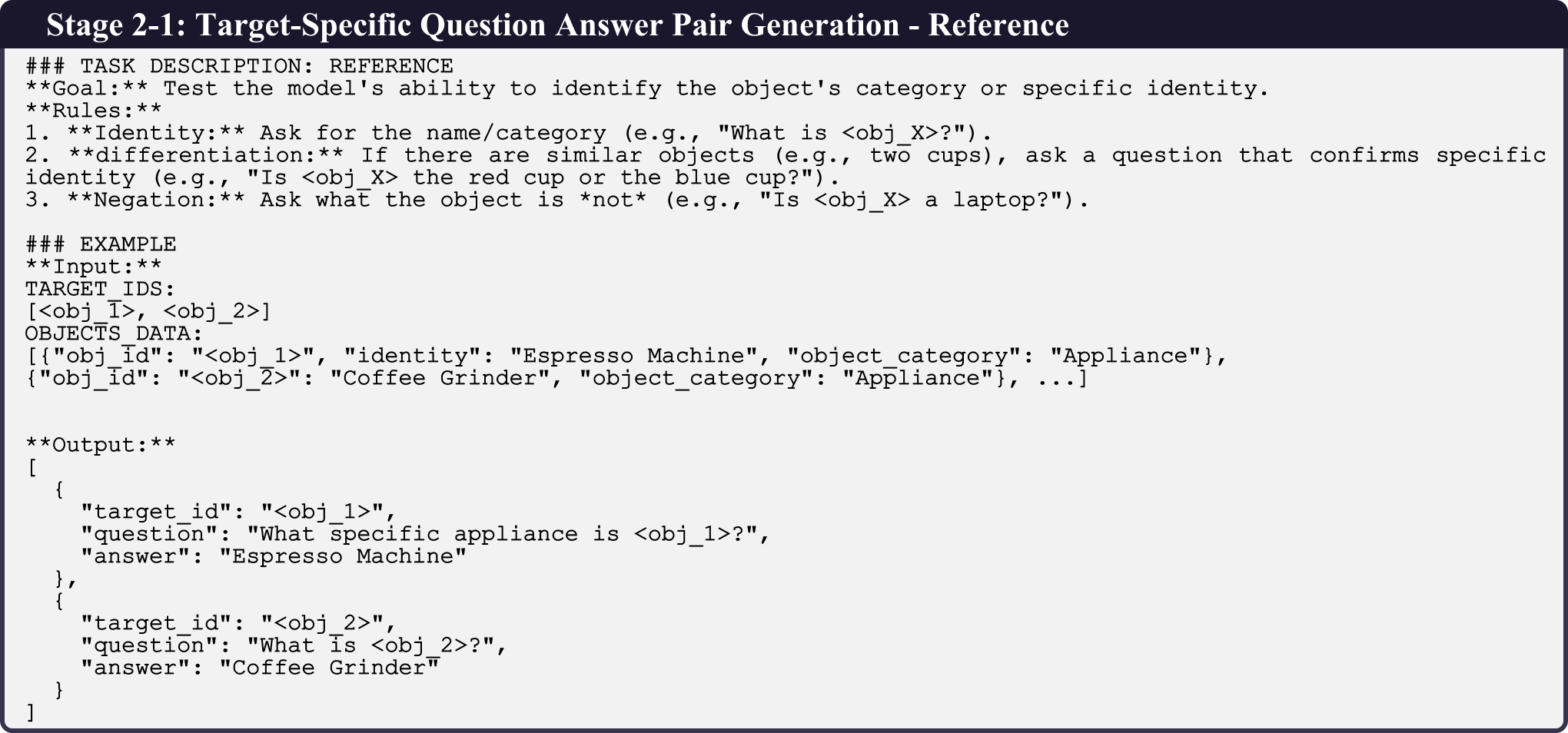}
    \vspace{-2em}
    \caption{\textbf{Task-specific prompt for Reference QA generation.}
    This task directs the model to generate questions that require identifying the specific description or category of the pointed-at object (e.g., ``What is this?").}
    \label{fig:supp_prompt_stage2_1_reference}
\end{figure*}

\begin{figure*}[t]
    \centering
    \vspace{-2em}
    \includegraphics[width=\linewidth]{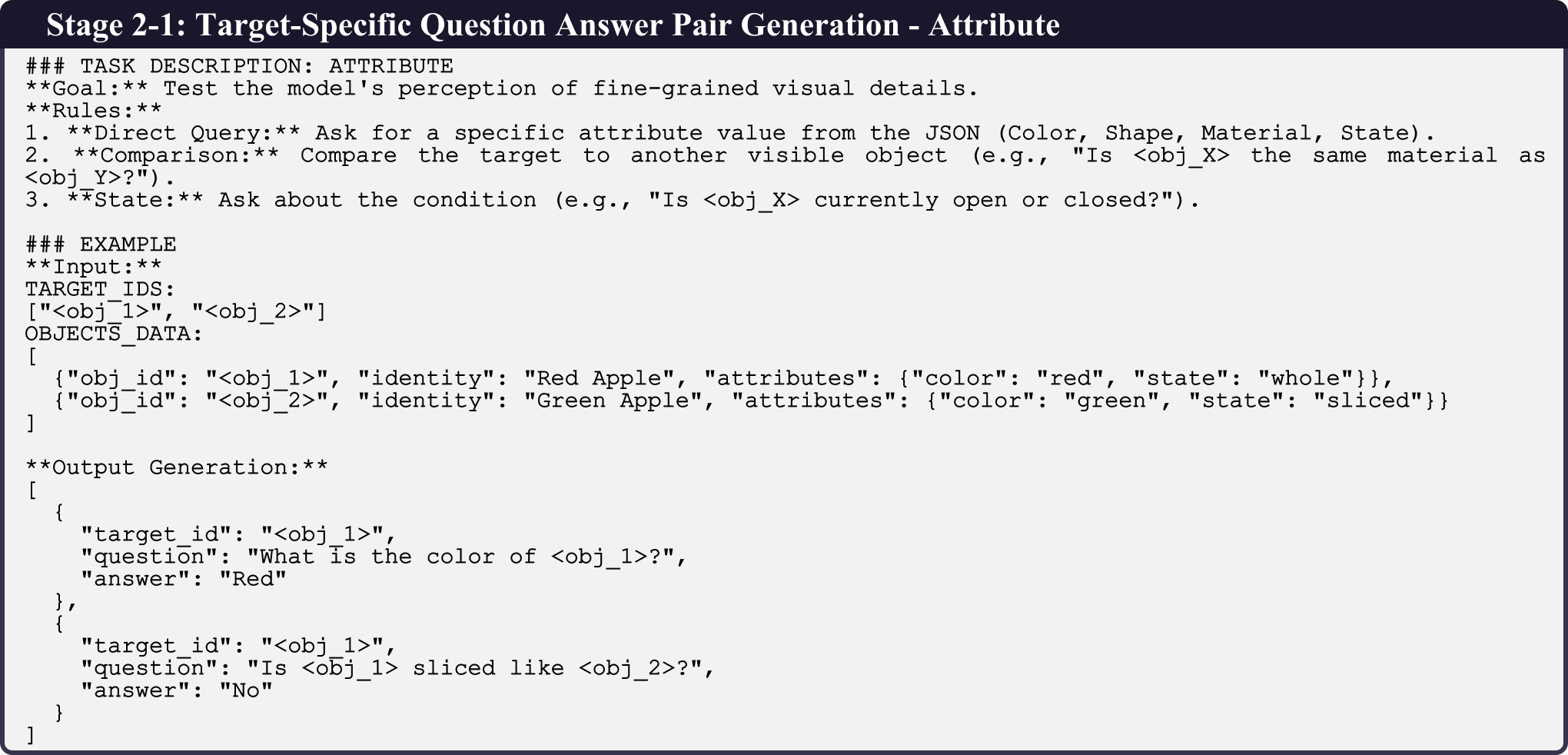}
    \vspace{-2em}
    \caption{\textbf{Task-specific prompt for Attribute QA generation.} This module focuses on fine-grained visual details, instructing the model to query properties such as color, shape, material, or state (e.g., ``Is this sliced?").}
    \label{fig:supp_prompt_stage2_1_attribute}
\end{figure*}

\begin{figure*}[t]
    \centering
    \vspace{-2em}
    \includegraphics[width=\linewidth]{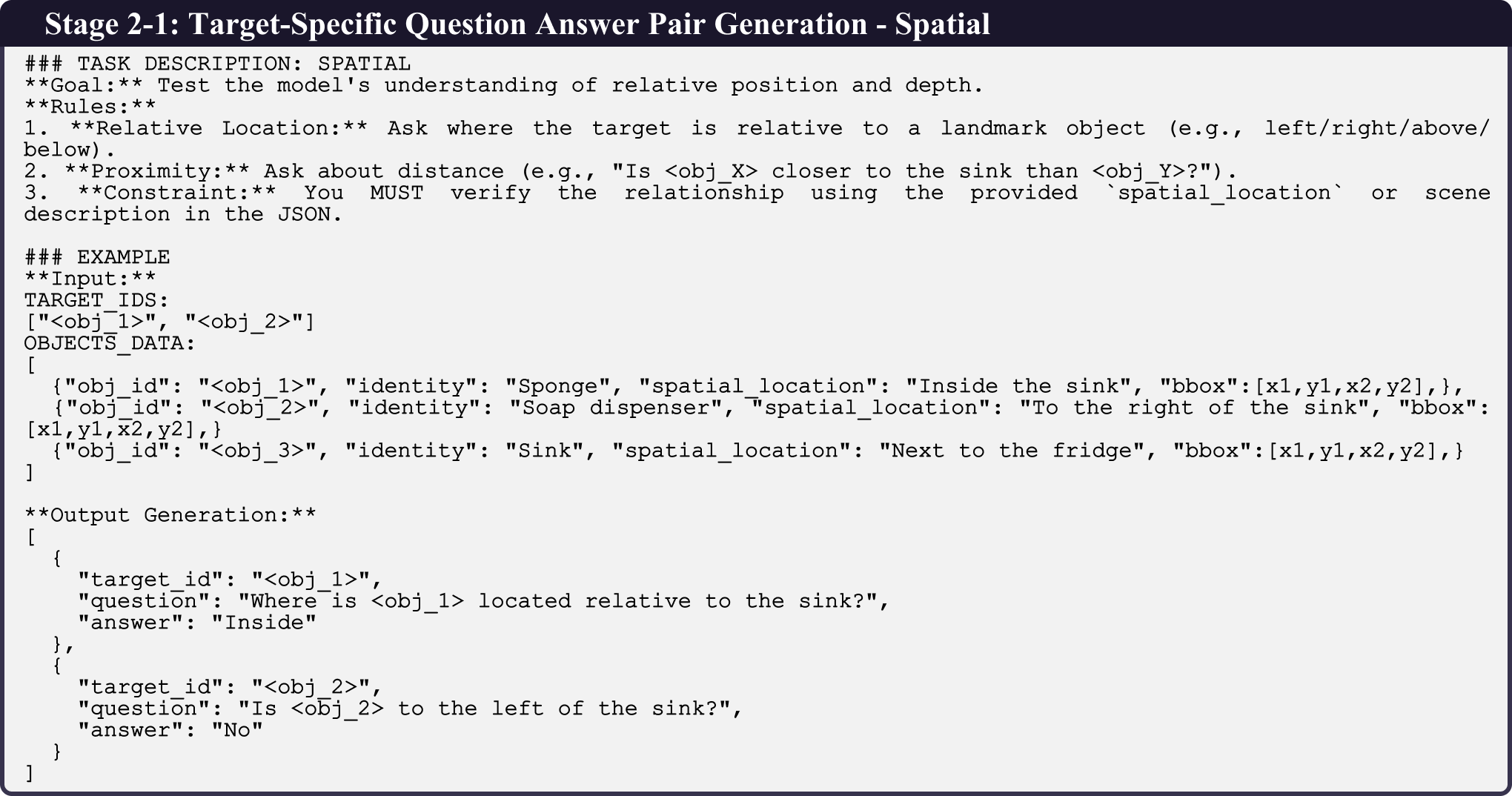}
    \vspace{-2em}
    \caption{\textbf{Task-specific prompt for Spatial QA generation.} This task generates questions regarding the relative position or depth of the target object compared to other landmarks in the scene (e.g., ``Is this to the left of the sink?").}
    \label{fig:supp_prompt_stage2_1_spatial}
\end{figure*}

\begin{figure*}[t]
    \centering
    \includegraphics[width=\linewidth]{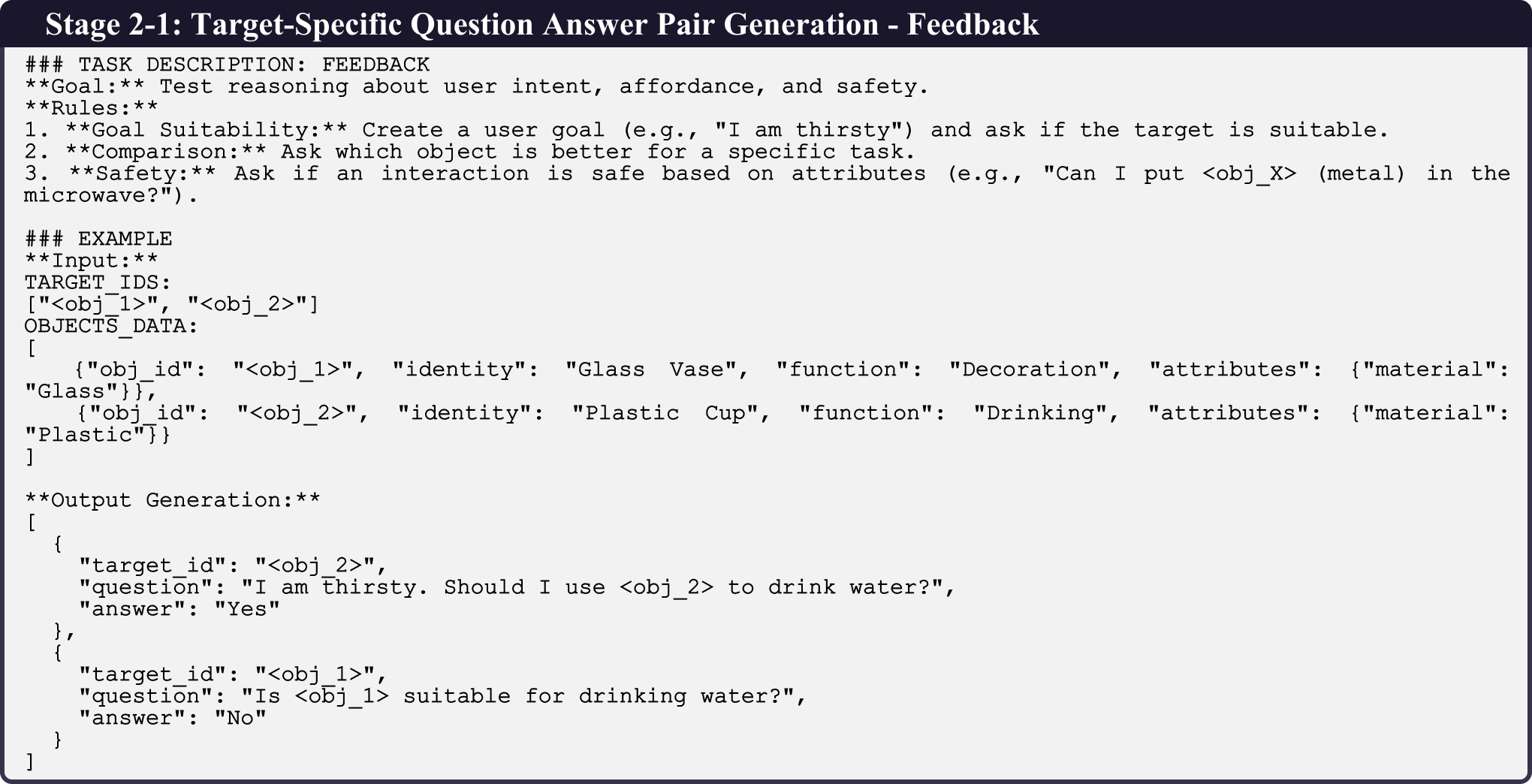}
    \vspace{-2em}
    \caption{\textbf{Task-specific prompt for Feedback QA generation.} This task tests reasoning about user intent and affordance, generating questions about whether a specific object is suitable for a stated goal (e.g., ``I am thirsty. Can I drink this?").}
    \label{fig:supp_prompt_stage2_1_feedback}
\end{figure*}

\begin{figure*}[t]
    \centering
    \includegraphics[width=\linewidth]{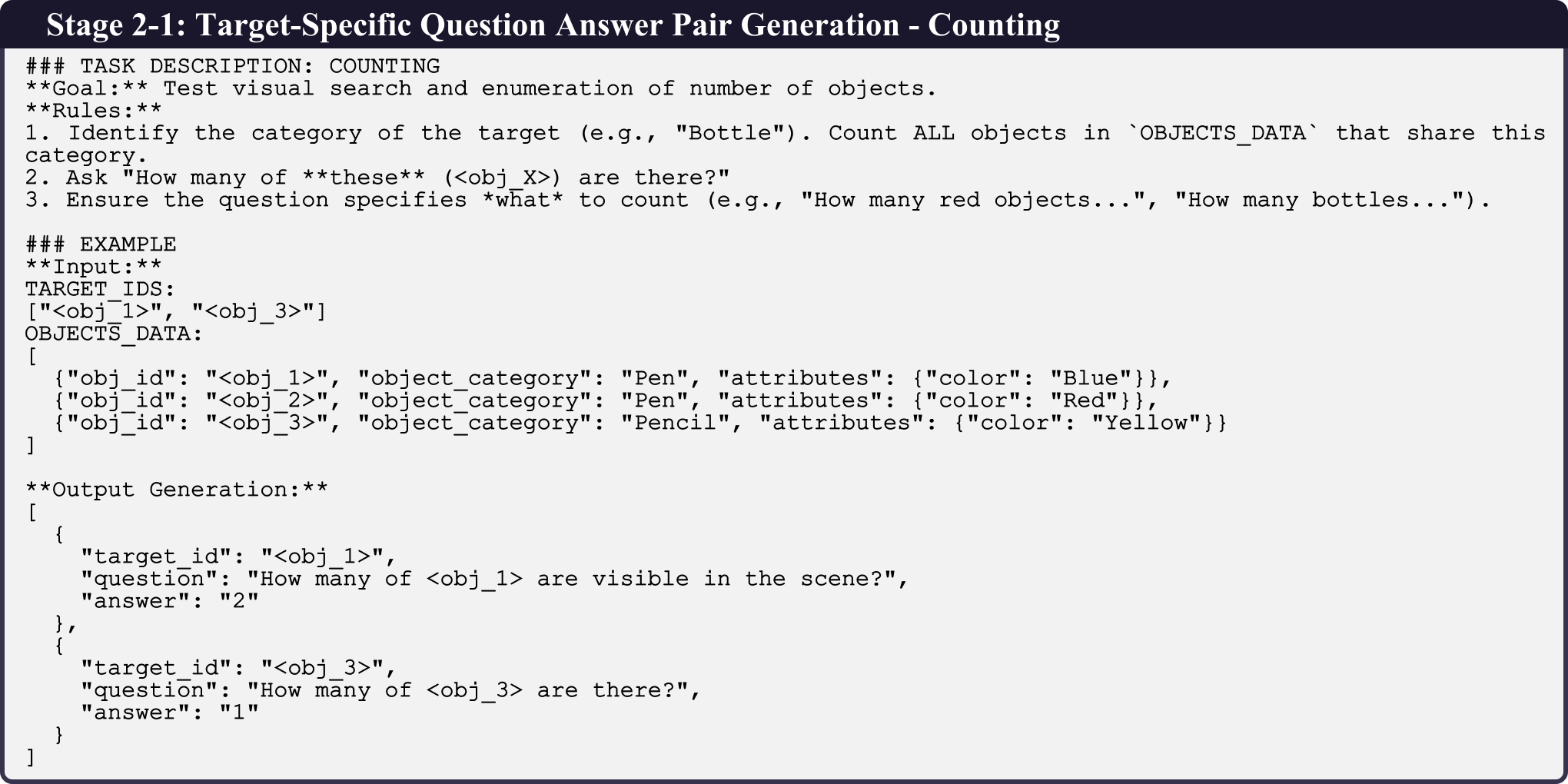}
    \vspace{-2em}
    \caption{\textbf{Task-specific prompt for Counting QA generation.}
    This task tests visual search capabilities by asking the model to enumerate instances of the pointed-at object category visible in the scene.}
    \label{fig:supp_prompt_stage2_1_counting}
\end{figure*}

\begin{figure*}[t]
    \centering
    \includegraphics[width=\linewidth]{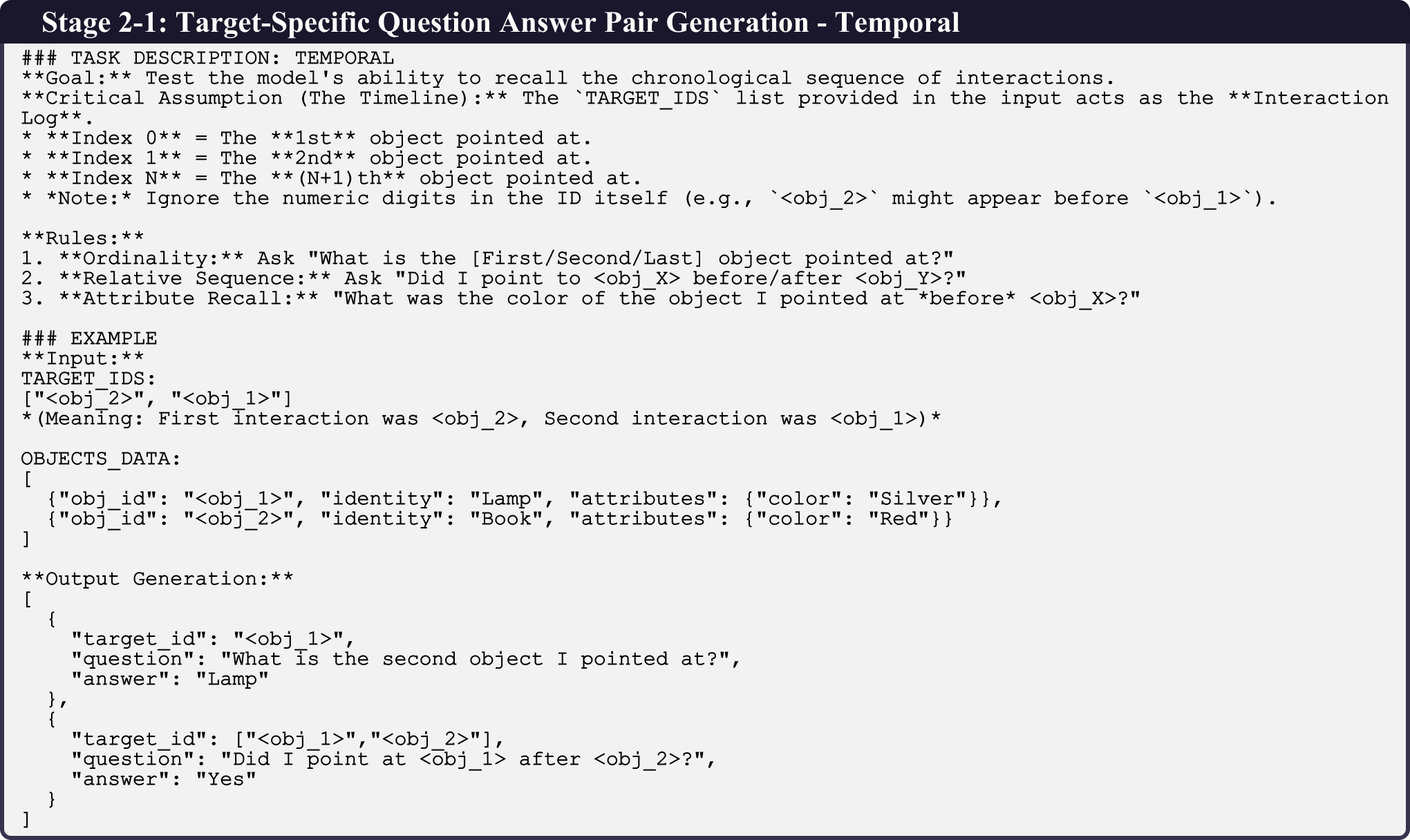}
    \vspace{-2em}
    \caption{\textbf{Task-specific prompt for Temporal QA generation.}
    This task challenges the model's ability to recall the chronological sequence of pointing gestures (e.g., ``What is the second object I pointed at?").}
    \label{fig:supp_prompt_stage2_1_temporal}
\end{figure*}

\noindent{\textbf{Stage 2-1: Question-answer pair generation.}} 
First, we feed the scene JSON from Stage 1 and a list of target object IDs into the MLLM. We use a modular prompting approach: a general instruction template (Fig.~\ref{fig:supp_prompt_stage2_1_general}) is combined with specific \textbf{Task Modules} depending on the question category. The General Template enforces that the answer must be derived \textit{exclusively} from the provided scene JSON to prevent hallucinations. The Task Modules contain specific logic for our six tasks: Reference (Fig.~\ref{fig:supp_prompt_stage2_1_reference}), Attribute (Fig.~\ref{fig:supp_prompt_stage2_1_attribute}), Spatial (Fig.~\ref{fig:supp_prompt_stage2_1_spatial}), Feedback (Fig.~\ref{fig:supp_prompt_stage2_1_feedback}), Counting (Fig.~\ref{fig:supp_prompt_stage2_1_counting}), and Temporal (Fig.~\ref{fig:supp_prompt_stage2_1_temporal}).



\begin{figure*}[t]
    \centering
    \vspace{-2em}
    \includegraphics[width=\linewidth]{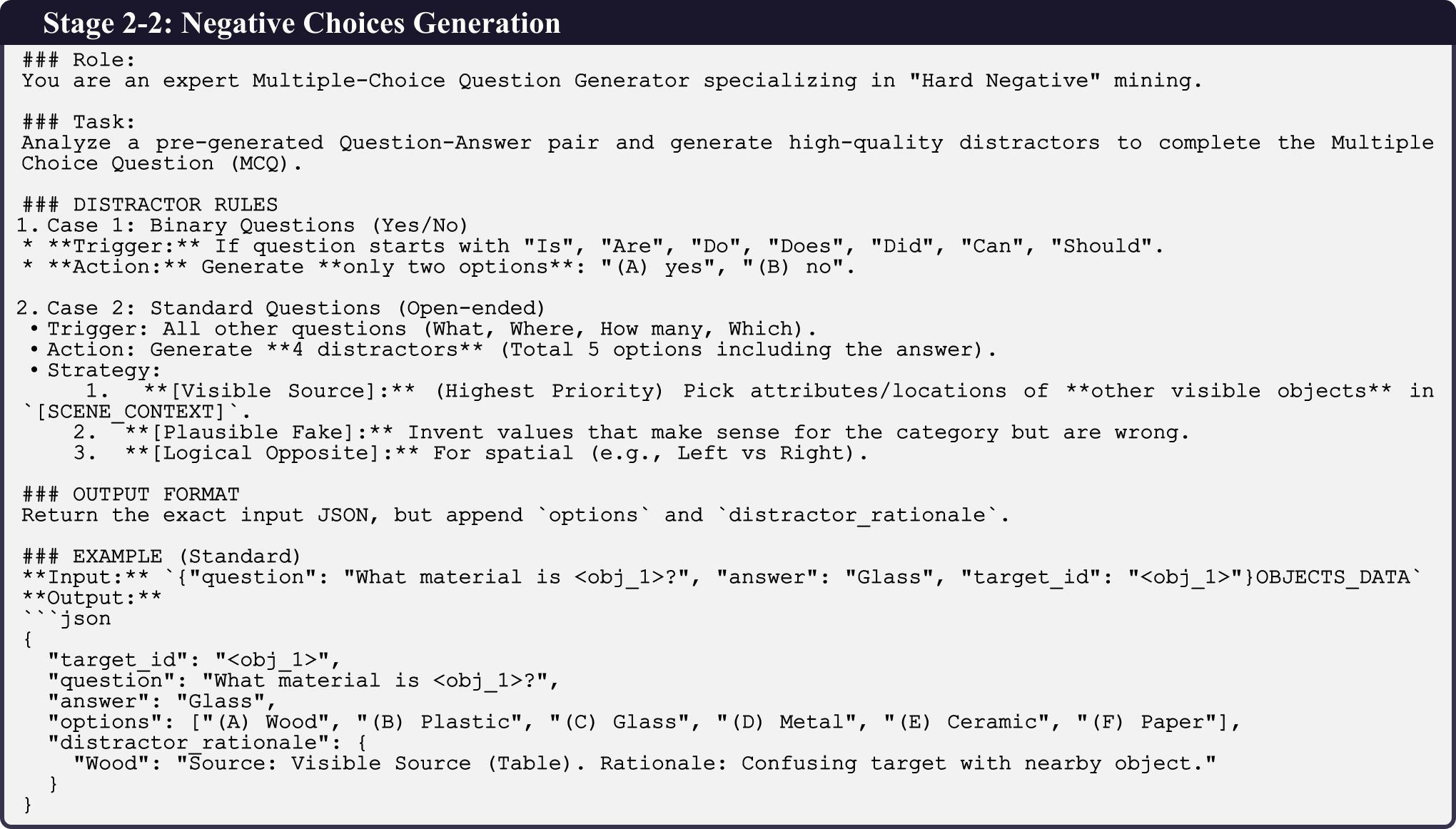}
    \vspace{-2em}
    \caption{\textbf{Prompt used for negative choices generation (Stage 2-2).}
    Instructions for generating negative choices.
    The model is prompted to select attributes from neighboring objects or plausible but incorrect properties to create challenging multiple-choice options.}
    \label{fig:supp_prompt_stage2_2_neg}
\end{figure*}

\noindent{\textbf{Stage 2-2: Negative choices generation.}} 
To convert the QA pairs into challenging Multiple-Choice Questions (MCQs), we prompt the model to generate distractors based on the specific question type, as detailed in Fig.~\ref{fig:supp_prompt_stage2_2_neg}.
For binary questions, the model is restricted to ``Yes/No'' options.
For open-ended questions, we instruct the model to generate four ``Hard Negatives'' using a prioritized strategy: (1) Visible Sources: Attributes or locations of \textit{neighboring} objects in the scene (e.g., if the target is red, select the color of a nearby blue cup). (2) Plausible Fakes: Attributes that are semantically valid for the category but visually incorrect. (3) Logical Opposites: For spatial or temporal relations (e.g., ``Left'' vs. ``Right'').

\begin{figure*}[t]
    \centering
    \vspace{-1em}
    \includegraphics[width=\linewidth]{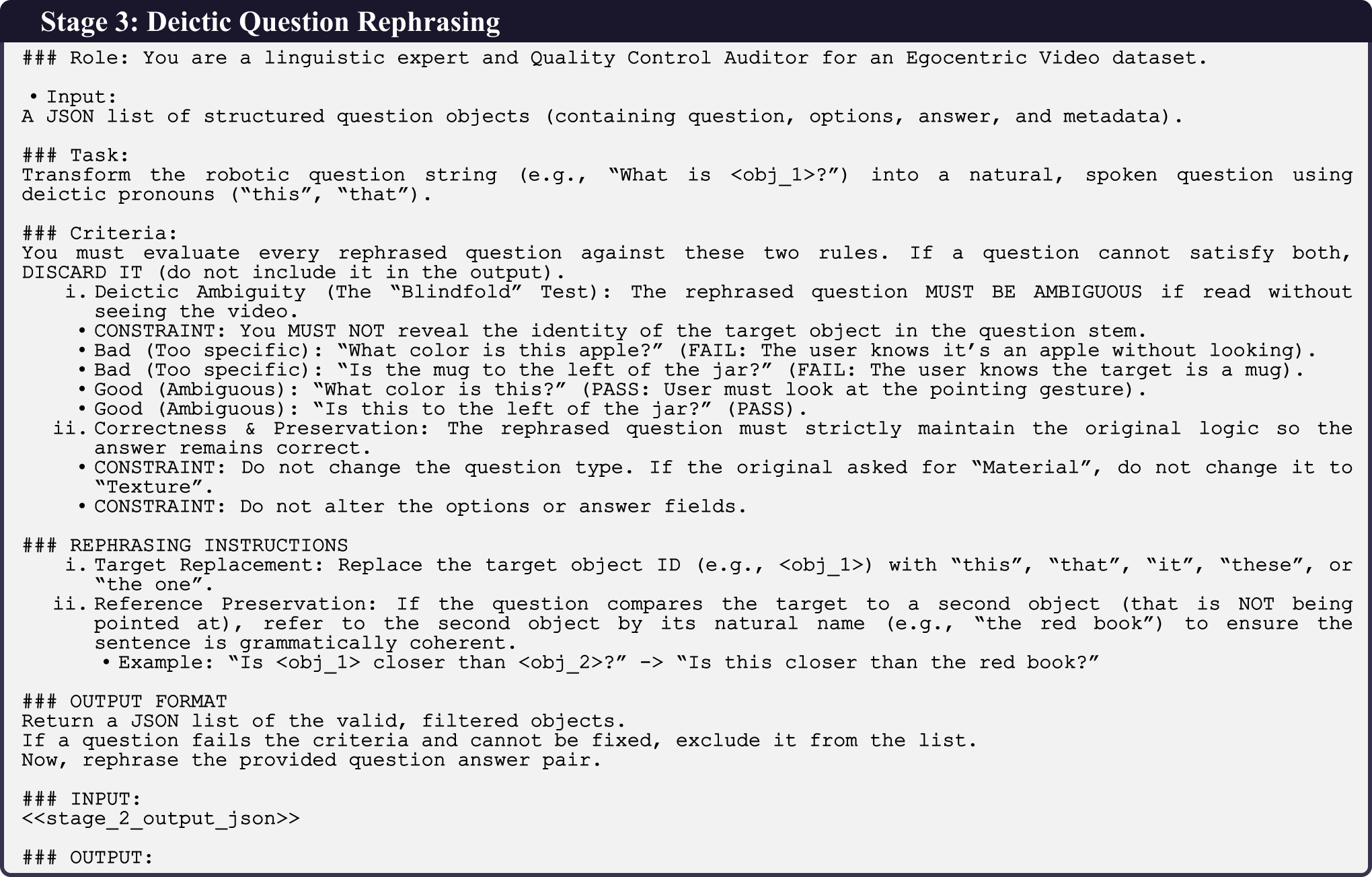}
    \vspace{-2em}
    \caption{\textbf{Prompt used for deictic question rephrasing (Stage 3).}
    The final prompt in the pipeline, where GPT-4o acts as a linguistic expert to convert structured queries into natural, spoken questions. It replaces explicit object names with deictic pronouns (``this", ``that") to ensure the question requires visual grounding to be answered.}
    \label{fig:supp_prompt_stage3_rephrase}
\end{figure*}

\noindent{\textbf{Stage 3: Deictic question rephrasing.}}
In the final stage, we prompt GPT-4o to transform the structured, robotic questions into natural, spoken egocentric queries (Fig.~\ref{fig:supp_prompt_stage3_rephrase}).
The prompt acts as both a linguistic rephraser and a quality control.
The model is instructed to replace target placeholders with deictic pronouns (``this'', ``that''), while preserving the names of reference objects (anchors) that are not being pointed at (e.g., ``Is \textbf{this} closer than the red book?'').
Here, we enforce that the rephrased question must be ambiguous if read without seeing the pointing gesture.

\begin{figure*}[t]
    \centering
    \vspace{-2em}
    \includegraphics[width=\linewidth]{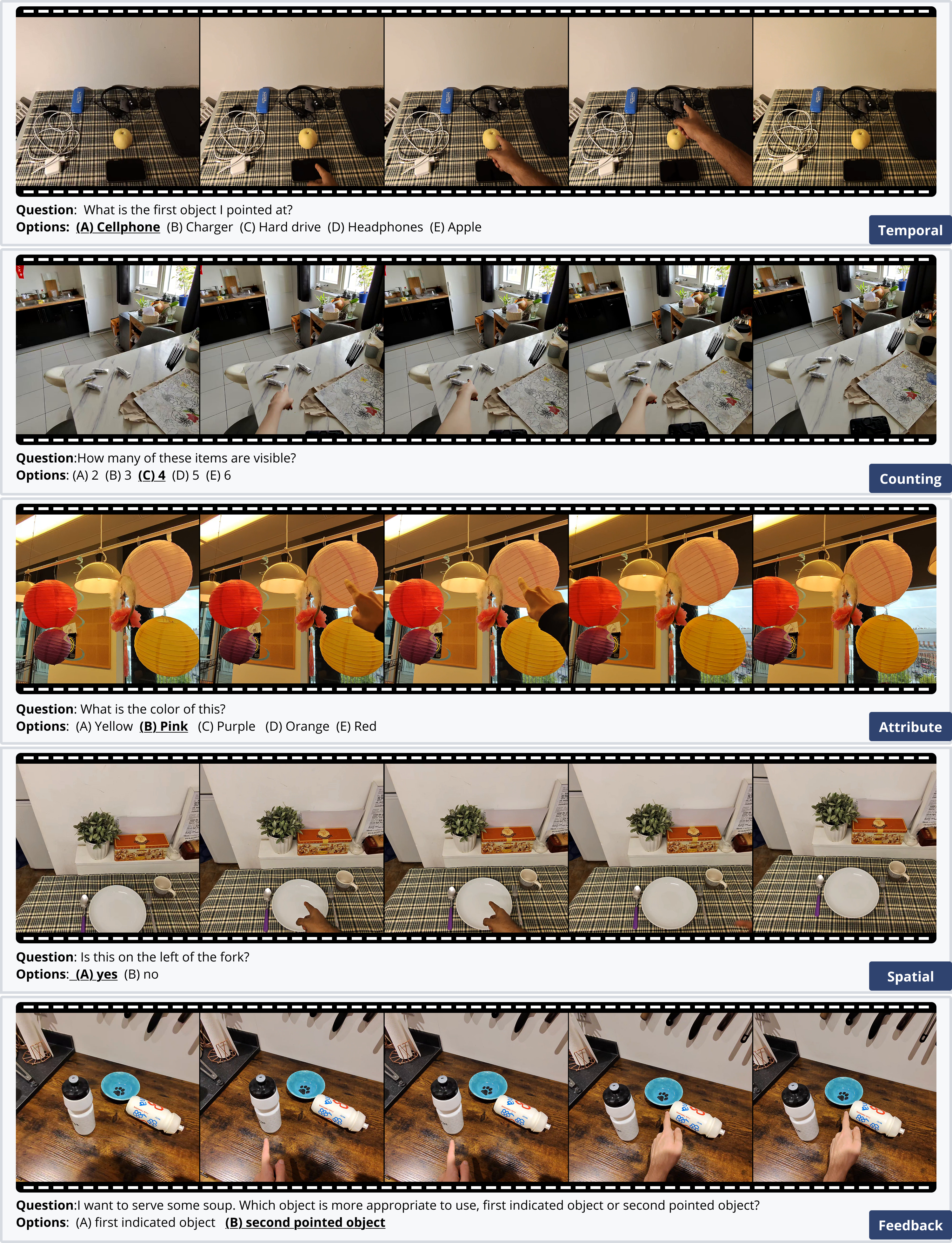}
    \vspace{-2em}
    \caption{\textbf{Examples from real-world indoor scenes.}
    Samples from the real-world split of the dataset, captured in indoor environments.
    Each row displays the sample video frames alongside the corresponding question-answer pair generated by our pipeline.
    The correct answer is bolded and underlined within the options.
    }
    \label{fig:supp_real_video_frames_indoor}
\end{figure*}

\begin{figure*}[t]
    \centering
    \vspace{-1em}
    \includegraphics[width=0.98\linewidth]{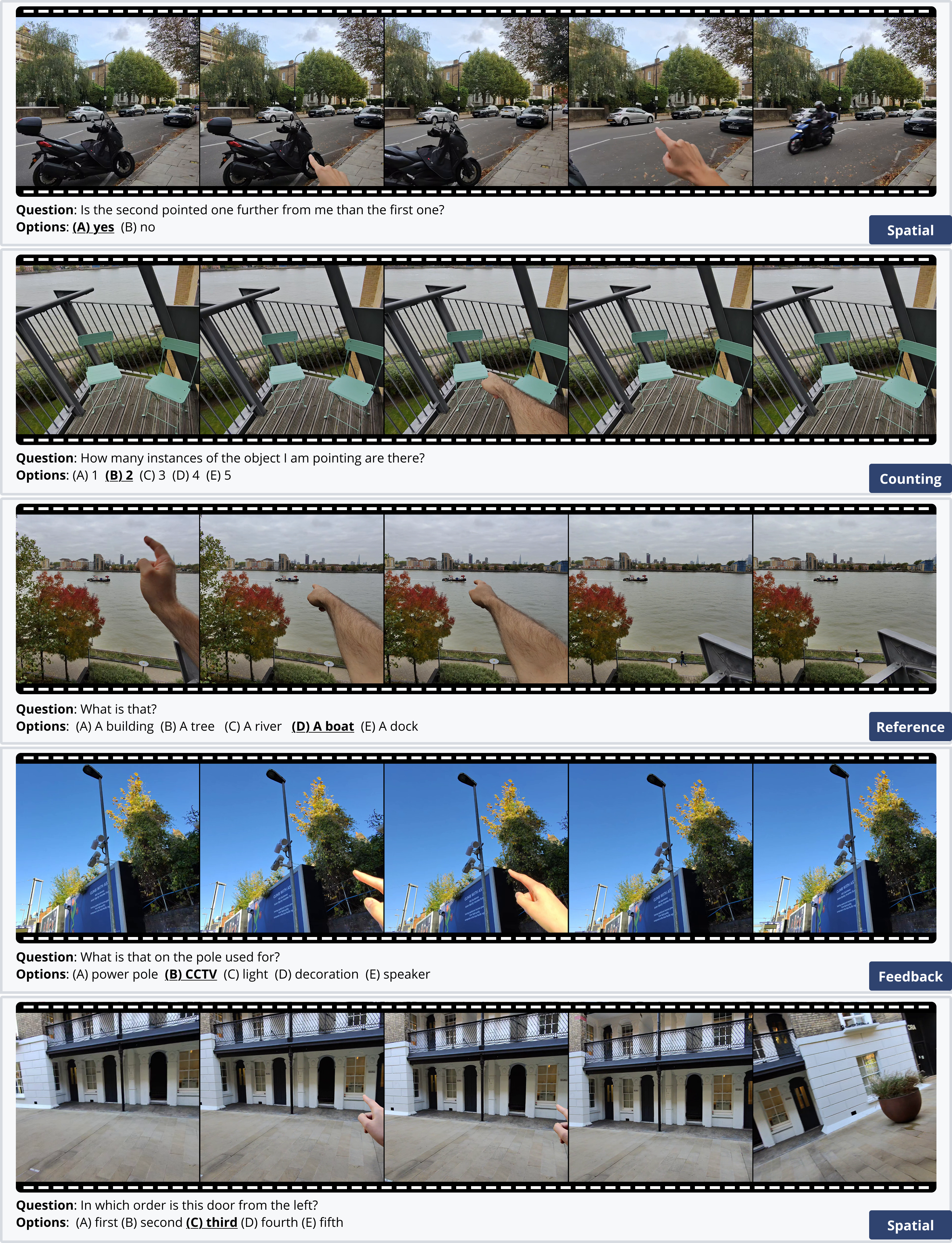}
    \vspace{-1em}
    \caption{\textbf{Examples from real-world outdoor scenes.}
    Samples from the real-world split of the dataset, captured in outdoor environments.
    Each row displays the sample video frames alongside the corresponding question-answer pair generated by our pipeline.
    The correct answer is bolded and underlined within the options.
    }
    \label{fig:supp_real_video_frames_outdoor}
\end{figure*}

\begin{figure*}[t]
    \centering
    \vspace{-1em}
        \includegraphics[width=\linewidth]{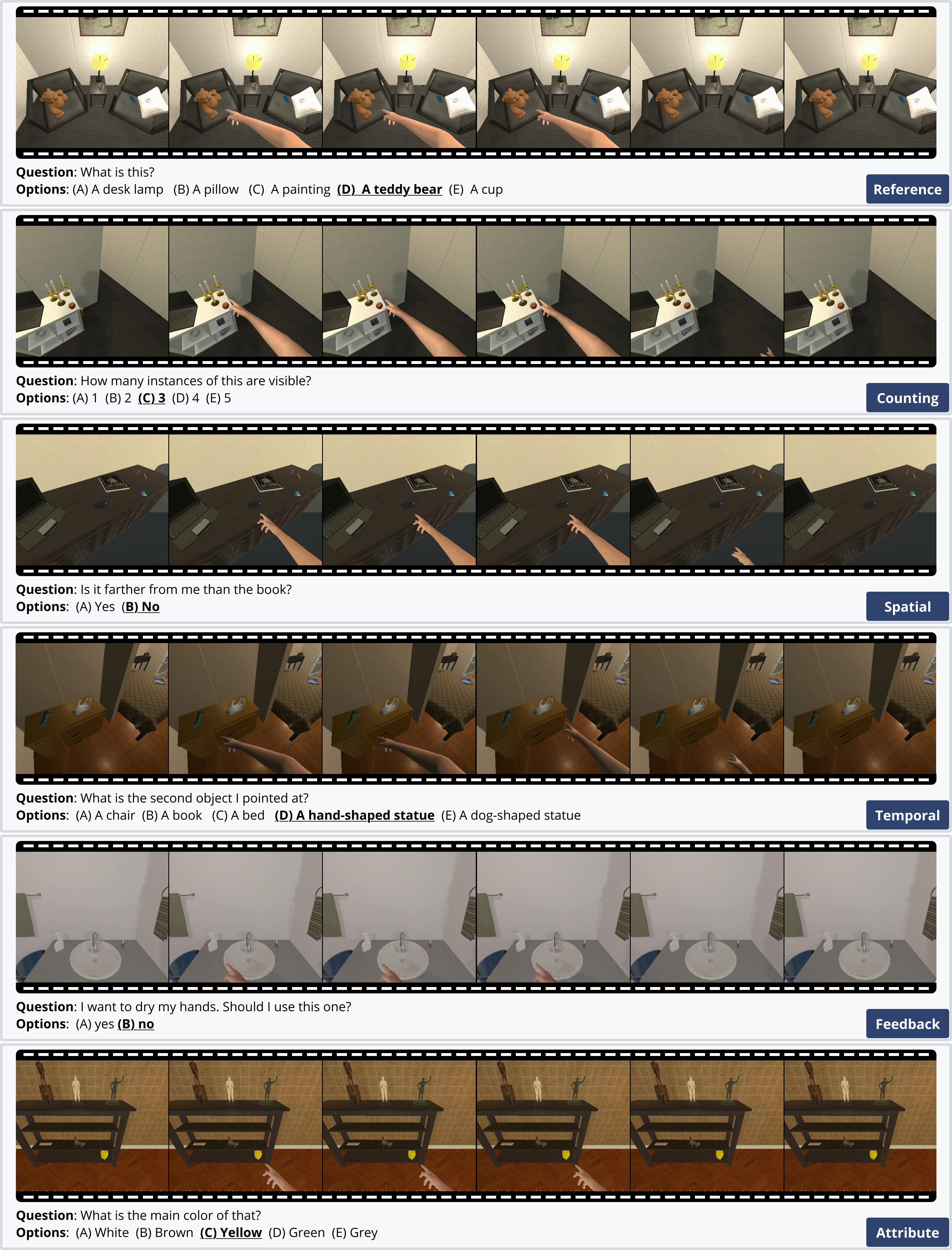}
    \vspace{-2em}
    \caption{\textbf{Examples from synthetic training data.}
        Samples from the large-scale synthetic training set generated via AI2-THOR.
        }
    \label{fig:supp_synth_train_qas}
\end{figure*}

\end{document}